\documentclass[10pt, a4paper]{article}

\usepackage[final]{lrec-coling2024} 

\usepackage{graphicx}
\usepackage{subcaption}
\usepackage{multirow}
\usepackage{makecell}
\usepackage{tabularx}
\usepackage{colortbl}
\usepackage{enumitem}
\usepackage{pgfplots}
\usepackage{csvsimple-l3}
\usepackage{siunitx}
\usepackage{fp}
\usepackage{ifthen}

\newcommand\perc[1]{\FPmul\result{#1}{100}\num[round-mode = places, round-precision = 2]{\result}}
\newcommand\std[1]{\textsuperscript{\color{gray}$\pm$\perc{#1}}}
\newcommand\stdstr[1]{\textsuperscript{\color{gray}$\pm$#1}}

\newcounter{notecounter}

\newcommand{\enoteson}{\long\gdef\enote##1##2{{
			\stepcounter{notecounter}
			\large\bf
			\hspace{100cm}\arabic{notecounter} $<<<$ ##1: ##2
			$>>>$\hspace{1cm}}}}
\enoteson

\definecolor{lightgray}{rgb}{.9,.9,.9}

\title{How to Solve Few-Shot Abusive Content Detection \\ Using the Data We Actually Have}

\name{Viktor Hangya and Alexander Fraser}

\address{Center for Information and Language Processing, LMU Munich, \\
        Munich Center for Machine Learning, \\
         \{hangyav, fraser\}@cis.lmu.de\\}

\abstract{
Due to the broad range of
social media
platforms,
the requirements of
abusive language detection systems are varied and ever-changing.
Already a large set of annotated corpora with different properties and label
sets were created,
such as hate
or misogyny detection,
but
the form and targets of abusive speech are constantly
evolving.
Since, the annotation of new corpora
is expensive,
in this work we leverage
datasets
we already have,
covering a wide range of tasks
related to abusive language detection.
Our goal is
to build
models cheaply
for a new
target label set and/or
language, using only a few training examples
of the target domain.
We propose a two-step approach: first we train our model in a multitask
fashion. We then carry out
few-shot adaptation to the target requirements.
Our experiments show that
using
already existing datasets
and
only
a few-shots
of the target task the performance of models
improve
both
monolingually
and across languages.
Our analysis also shows that our models acquire a general understanding of
abusive language, since they
improve the prediction of labels which are present only in the target dataset
and
can benefit from knowledge about labels which are not directly used for the
target task.
 \\ \newline  }

\begin{document}

\maketitleabstract

\section{Introduction}
\label{sec:intro}

The wide spread of social media allowed us to communicate and share our opinions
quickly
and conveniently.
However, it
gives
place to abusive content as well, which leaves some groups of
people vulnerable.
To push back abusive online content, various automated systems, and more
importantly datasets \cite{poletto2021resources},
were introduced covering various text genres such as
forum~\cite{de-gibert-etal-2018-hate}, Twitter~\cite{struss2019overview} or
Instagram posts~\cite{suryawanshi-etal-2020-multimodal} of various
languages~\cite{Vidgen2020}, user groups such as
women~\cite{fersini2018overviewEvalita} or
LGBTQ+~\cite{leite-etal-2020-toxic} and tasks including hate
speech~\cite{de-gibert-etal-2018-hate}, offensive
language~\cite{zampieri-etal-2019-predicting} or
toxicity~\cite{leite-etal-2020-toxic} detection, etc.

\begin{figure}[t]
    \centering
    \includegraphics[width=0.45\textwidth]{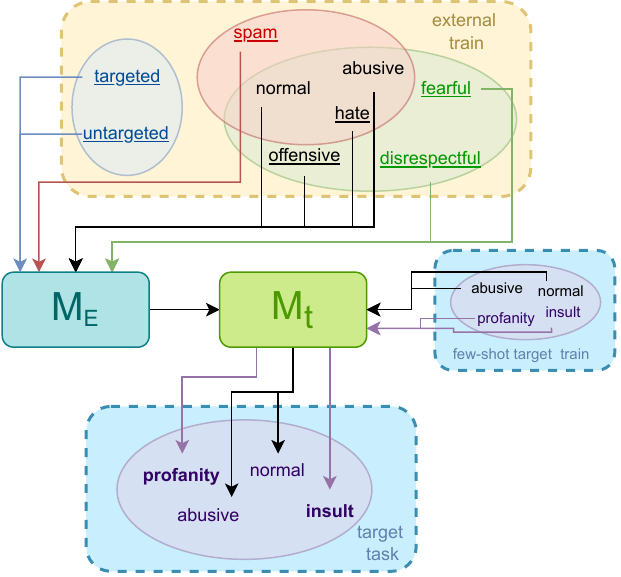}
    \caption{
    Two-step approach: $M_E$ is trained on the (external) datasets we already
    have, followed by
    its adaptation to the target task ($M_t$) with only
        a few-shots.
    Labels not directly used for the target task are \underline{underlined},
    target labels not contained in the external datasets are \textbf{bolded}.
    }
    \label{fig:example}
\end{figure}

On the other hand,
there is
constantly
a need to annotate new datasets supporting
previously unseen
target scenarios.
To reduce annotation costs,
related work leveraged transfer learning
to build systems across languages \cite{Ranasinghe2020} and domains
\cite{Glavas2021}.
But finding the right source datasets
is often
challenging, since the label sets
could differ.
To alleviate the problem, previous work manually altered
the label sets of the
source datasets in order to match them to the target requirements.
However, this
approach is problematic, because it
requires
expertise in abusive language datasets,
since i) the
rules developed by other researchers
for manual label matching are not reusable due
to the rapid change in the application requirements
and ii) the definition of the same label in some datasets could
conflict, e.g., the
offensive label of the \texttt{OLID} dataset includes profane language
\cite{zampieri-etal-2019-predicting}, while the same label does not in
\texttt{HASOC} \cite{Mandl2019}.
Thus, a precise understanding of abusive
language phenomena is required.
Additionally,
iii) novel fine-grained labels do not have alternatives to be transferred from.
Such fine-grained labels, e.g., related to a specific view of a given community
regarding an event, can be created on-the-fly as moderators or affected people
face them.
Thus, our goal is to eliminate the need for such rules and
make information transfer more flexible
with minimal target task annotations.

To this end,
we introduce a
method leveraging multiple already existing
(external)
datasets in order to
understand general abusive language,
allowing to build models cheaply for
the target requirements
without the need for manual dataset modifications.
As shown in Figure~\ref{fig:example}, different datasets can inform the model about
different types of abusive content.
Some classes
can directly be leveraged for
the target task due to their matching
label names and definitions.
Others,
such as labels with matching names but conflicting definitions or labels which
are not contained in the target dataset at all,
are leveraged only indirectly,
by contributing to the general abusive language awareness of the
model.
Our approach
consists
of two steps: jointly training a language model on
multiple external datasets using
prompt-learning \cite{schick-schutze-2021-exploiting}.
We then adapt the resulting
model to the target
requirements in the second step, using only
a few
samples
per label
from the
target task
(4-shots in the main experiments),
which could even be created on-the-fly.
Since the target task can contain unseen labels, i.e., labels which are not
contained in any of the external datasets,
or classes with conflicting label definitions, at least a few
annotated samples
are needed
for model specialization.

We test our method on various tasks (e.g., hate,
abuse and
misogyny detection or target identification) in both monolingual and
cross-lingual (English$\rightarrow$German,
$\rightarrow$Italian, $\rightarrow$Brazilian~Portuguese, and
$\rightarrow$Hindi) setups.
Additionally, our datasets cover multiple platforms including longer forum posts
and
shorter Twitter messages.
Our experiments show improved performance when training using the external
datasets compared to various baselines, including both monolingual and
cross-lingual settings, on both binary
and fine-grained test sets.
We find
that
even
unseen target
labels
are
improved due to the better general abusive language understanding of
our models.
Our ablation study shows that external-only labels
(labels which do not occur in the label set of the task we are carrying out)
improve performance, showing that they contribute to general understanding as well.
We experiment
with different target data sizes and
find
that although our
approach is more beneficial at lower sizes,
when more data is available it is also effective.
Finally, we perform
model diagnostics using \textsc{HateCheck}
\cite{rottger-etal-2021-hatecheck}, further supporting our claim of better
general abusive language understanding.
Our contributions are the
following:\footnote{
Our code is publicly available at: \\ \url{https://cistern.cis.lmu.de/multi_hs}
}
\begin{itemize}

        \item a multi-dataset training (MDT) approach,
                using
            prompt-learning
    based fine-tuning,
    for an efficient few-shot training
    which supports the ever-changing nature of abusive language detection,

        \item applicability across languages and text genres to support a wide range
    of target tasks cheaply,

        \item
comprehensive
        analyses for a better understanding of
        model behavior.
                                    \end{itemize}

\section{Related Work}
\label{sec:rel_work}

To alleviate the issue
of missing datasets for a given target task, previous
work leveraged transfer learning techniques.
\citet{Ranasinghe2020} built hate speech classifiers for Hindi, Spanish
and Bengali by relying only on an English training dataset,
while \citet{Glavas2021} followed a similar approach for cross-domain
experiments.
They made
the train and test corpora compatible
using rule based label adaptation which, as discussed above, is often difficult.
In this work, we eliminate this step and use external datasets without any
modifications.
Furthermore, \citet{Wiegand2018} showed that by adding seemingly similar English
samples to a small amount of German training data
the results decreased,
while \citet{nozza-2021-exposing} found
that
in zero-shot cross-lingual models
language specific interjections are often misinterpreted
leading
to errors.
These results indicate
that selecting the right source dataset is not straight
forward
(and perhaps impossible in some cases).
In this work, we leverage multiple external datasets
for a robust abusive language understanding,
and use
a few-shots
from the target dataset to
specialize our models to the target domain.
Similarly, \citet{rottger-etal-2022-data} argues for the use of a small amount
of target language training samples in order to extend hate
detection to
multiple languages.
However, they focus on compatible binary datasets, while our approach is
compatible with fine-grained tasks with unseen labels as well.

In contrast to
transfer learning,
the goal of multitask learning (MTL)
is to build a
shared model using various tasks in order to improve the
performance on all of them,
by exploiting common information in some tasks, and to perform multiple tasks
with a single model \cite{caruana1997multitask}.
\citet{pmlr-v97-stickland19a} proposed an MTL method based on pre-trained
language models by introducing task specific parameters in each layer.
Due to negative task interference however, single task models perform best in
many cases.
To mitigate the issue
of task interference,
\citet{pfeiffer-etal-2021-adapterfusion} used adapters
\cite{pmlr-v97-houlsby19a} in
a
multitask setting, showing that fusing information
learned by task specific adapters can further boost the performance on a target
task.
However, MTL is not able to induce useful task dependencies when
an imbalanced set, in terms of dataset sizes, is available.
In contrast,
a set of auxiliary tasks were used
to improve the performance on a single target task in
\cite{watanabe-etal-2022-auxiliary}.
Similarly, \citet{MEHMOOD2020848} perform a final training step on the target
biomedical NER task after MTL,
while \citet{Kapil2020} combine various abusive language related datasets.
However, these methods rely on a large set of training data for the target task
in order to improve the performance,
which is
often
unavailable in the ever-changing field of abusive language detection.
\citet{magnossao2023mitigating} proposed to use task embeddings in order to
reduce negative transfer in MTL, but they only consider binary abusive language
detection tasks.
In comparison,
we
use a wide range of tasks with heterogeneous label sets,
including external-only
labels, when just a little target data is available, and we show that
this
leads to
a better general abusive language understanding which positively impacts even
unseen target labels.

Our approach is also related to meta-learning \cite{hospedales2021meta}, where
the goal is to build a general model that is cheaply adaptable to a target
task that is unknown at the time of meta-learning.
In contrast, our goal is to build a classifier for a resource poor task that is
known when training the full model, by leveraging already existing highly
related datasets without strong expertise in abusive language phenomena.
Additionally,
\citet{wang2021bridging} showed that
meta-learning
has similar performance
to MTL,
thus
we only use MTL as
a baseline for simplicity.

\section{Approach}
\label{sec:approach}

We consider two sets of training corpora
in our multi-dataset training (MDT) approach: external datasets
($D_E=\{D_{e_i}: i=1..N\}$)
which are not directly related to the target task
and the target dataset ($D_t$) which is
the target task
for which we aim to build a classifier.
The former are off-the-shelf datasets created for other tasks
and/or
languages
containing a few thousands or
sometimes
tens of thousands of samples.
In contrast, since our main goal is to reduce the costs of building systems for
novel target tasks,
$D_t$
contains only a few samples,
4-shots per label in our main experiments.
MDT builds
abusive
language classifiers in two steps (Figure~\ref{fig:example}).
First, we train a single model by fine-tuning a pre-trained LM ($M_0$) using
only the external datasets in order to learn general abusive language
understanding
(resulting in $M_E$), which we adapt to the specificities of the target task
in the second step (resulting in $M_t$).
In contrast to MTL or meta-learning, where the final model supports multiple tasks,
our
final
models ($M_t$) are built for a single target task.
This imitates the use cases of social media platforms which need to build a
specialized model supporting their own specific requirements.
First, we
discuss the used prompt-learning technique \cite{schick-schutze-2021-exploiting},
followed by the introduction of
the two-step training.

\subsection{Prompt-Learning}
\label{sec:prompt_learning}

Prompt-learning was shown to be effective when only a small training set is
available.
Instead of using classification heads on top of pre-trained LMs, it aims to
solve the task at hand using text generation.
Depending on the used LM architecture, various techniques exist.
We rely on encoder-only LMs in our experiments, thus use the method proposed by
\citet{schick-schutze-2021-exploiting}, which
employs the
masked language modeling task (MLM) to perform text classification.
Using pattern-verbalizer-pairs (PVPs) an input sentence is first transformed using
the pattern, e.g.,
\emph{I'll kill you.} $\rightarrow$ \emph{I'll kill you. It was [MASK]},
and the task is to
output the probability distribution over the vocabulary items at the masked
position.
Finally, the verbalizer maps the
highest probability token, out of a set of valid tokens
(see below),
to labels of a given dataset,
e.g., \emph{threatening $\rightarrow$ \texttt{threat}} or \emph{neutral
$\rightarrow$ \texttt{normal}}.
During
fine-tuning
the model is
trained to predict the token associated with the correct input label,
using the MLM objective.

In our multi-dataset setup, we define PVPs for all external datasets and the
target dataset separately ($PVP_E = \{PVP_{e_i}: i=1..N\}$ and $PVP_t$).
Having a dedicated
pattern and verbalizer
for each dataset
makes our approach easy to be
specialized for
each dataset,
and at the same time easy to use,
since
no single verbalizer
handling all the labels
is
required.
We defined only two different patterns: one for target detection datasets
(\texttt{X} $\rightarrow$ \texttt{X} It was targeted at \texttt{{[}MASK{]}},
where \texttt{X} is the input text)
and another for the rest
(\texttt{X} $\rightarrow$ \texttt{X} It was \texttt{{[}MASK{]}}).
On the other hand, the used verbalizers are specific for the label set of each
dataset, which can be defined easily in general,
i.e., we used 1-to-1 token to label mapping
in most cases.
We refer to Table~\ref{tab:datasets} of the Appendix for more details about the
exact patterns and verbalizers for each used dataset.

\subsection{Multi-Dataset Training}
\label{sec:multitraining}

\paragraph{Step 1: General Model Training ($M_0 \rightarrow M_E$)}
In each step of the training process we randomly select an external dataset
$D_{e_i}$ and a batch of samples from it.
Other than the shared model core, i.e., the pre-trained LM,
we use the PVP related to $D_{e_i}$
for the forward-backward pass.
For each dataset $D_{e_i}$ we use cross-entropy loss as the objective
function $L_{e_i}$ to update the model.
We run this process until convergence.

\paragraph{Step 2: Model Specialization ($M_E \rightarrow M_t$)}
In order to adapt $M_E$ to the target task,
we simply continue training it on $D_t$ by
using the $D_t$ specific PVP.
Similarly to the above, we use cross-entropy loss $L_t$ to update the model until
convergence.
As shown by our experiments, the general abusive language understanding learned
by $M_E$ helps this step to build a better model using just a few training
samples.

\section{Experimental Setup}
\label{sec:exp_setup}

\subsection{Datasets}
\label{sec:sec:datasets}

We selected a wide range of datasets for our experiments, covering various
abusive language detection tasks, languages and text genres.
We give a short overview in the following and further details, such as
number of samples,
exact PVPs, etc., in
Table~\ref{tab:datasets} of the Appendix.

\paragraph{AMI}
was created
for the \emph{Evalita 2018 shared task on Automatic Misogyny
Identification} \cite{fersini2018overviewEvalita}, containing English and
Italian tweets.
We use both the binary and fine-grained misogyny labels as well as the target
identification labels.

\paragraph{GermEval}
was introduced for the shared task on the \emph{Identification of Offensive
Language} in German tweets \cite{struss2019overview}.
We used both binary and fine-grained label sets.

\paragraph{HASOC}
The shared task on \emph{Hate Speech and Offensive Content Identification}
\cite{Mandl2019} introduces datasets for English, German and Hindi containing
Twitter and Facebook posts.
We used its fine-grained abuse and target identification labels.

\paragraph{HatEval}
was built for \emph{SemEval 2019 Task 5} about the detection of hate
speech against immigrants and women in Spanish and English Twitter
messages \cite{basile-etal-2019-semeval}.
We used its binary hate speech and target identification label sets.

\paragraph{LSA}
is a large scale fine-grained abusive dataset of English Tweets
\cite{Founta2018}.

\paragraph{MLMA}
\citet{ousidhoum-etal-2019-multilingual} introduced a multilingual and
multi-aspect hate speech dataset of English, French and Arabic Tweets.
We leveraged the fine-grained hostility labels in English.

\paragraph{OLID}
The Offensive Language Identification Dataset contains English tweets annotated
with offensive labels on three layers
\cite{zampieri-etal-2019-predicting}.
We used its binary offensive text and target identification subsets.

\paragraph{SRW}
is an English Twitter
set created for sexism and racism detection
\cite{waseem-hovy-2016-hateful}.

\paragraph{Stormfront}
was created for hate speech detection containing English forum posts from the
\emph{Stormfront} white supremacist forum \cite{de-gibert-etal-2018-hate}.
It is annotated with binary labels.

\paragraph{ToLD-Br}
is a Brazilian Portuguese Twitter dataset annotated for toxicity detection
\cite{leite-etal-2020-toxic}.
We used its fine-grained label set containing a wide range of labels, including
misogy\-ny.

\subsection{Multi-Dataset Setup}
\label{sec:sec:setus}

\begin{table}[t]
\centering
\renewcommand\tabularxcolumn[1]{m{#1}}\resizebox{\columnwidth}{!}{\begin{tabularx}{.68\textwidth}{>{\hsize=0.5\hsize}X>{\hsize=0.40\hsize\raggedright\arraybackslash}X}
Dataset                             & Labels                                                          \\ \hline
\multicolumn{2}{c}{External datasets ($D_E$)}                                                          \\
\hline
\hline
AMI binary misogyny En              & misogyny, normal                                                \\
\rowcolor{lightgray}
AMI fine-grained misogyny En        & stereotype, dominance, \underline{derailing}, sexual\_harassment, discredit \\
AMI binary target En                & active, passive                                                 \\
\rowcolor{lightgray}
HASOC fine-grained abusive En       & hate, offensive, profanity                                      \\
HASOC binary target En              & targeted, \underline{untargeted}                                            \\
\rowcolor{lightgray}
HatEval binary target En            & individual, group                                               \\
LSA fine-grained abusive En                 & abusive, hateful, \underline{spam}, normal                                  \\
\rowcolor{lightgray}
MLMA fine-grained hostility En      & abusive, hateful, offensive,
\underline{disrespectful}, \underline{fearful}, normal     \\
SRW fine-grained abusive En                 & sexism, racism, normal                                          \\
\hline
\multicolumn{2}{c}{Target datasets ($D_t$)}                                                            \\
\hline
\hline
HASOC fine-grained abusive En & hate, offensive, \textbf{profanity}                                      \\
\rowcolor{lightgray}
HASOC fine-grained abusive Hi & hate, offensive, \textbf{profanity}                                      \\
HASOC fine-grained abusive De & hate, offensive, \textbf{profanity}                                      \\
\rowcolor{lightgray}
GermEval fine-grained offensive De  & profanity, \textbf{insult}, abusive, normal                              \\
ToLD-Br fine-grained toxicity Pt-Br & \textbf{LGBTQ+phobia}, \textbf{obscene},
\textbf{insult}, racism,
misogyny, \textbf{xenophobia}, normal                                                  \\
\rowcolor{lightgray}
OLID fine-grained target En         & individual, group, \textbf{other}                                        \\
Stormfront binary hate En           & hate, normal                                                    \\
\rowcolor{lightgray}
HatEval binary hate En           & hateful, normal                                                 \\
HatEval binary hate Es           & hateful, normal                                                 \\
\rowcolor{lightgray}
OLID binary offensive En            & offensive, normal                                               \\
GermEval binary offensive De        & offensive, normal                                               \\
\rowcolor{lightgray}
AMI binary misogyny En           & misogyny, normal \\
AMI binary misogyny It           & misogyny, normal
\end{tabularx}}
\caption{
        Multi-dataset setup including external ($D_E$) and target ($D_t$) datasets.
                                                                We consider similarly defined but differently named
        labels to be the same, such as hate and hateful, sexism and misogyny or
        individual and active.
                        We bold \textbf{unseen} labels and underline labels \underline{which
        aren't used} in any target dataset.
        We remove external datasets from $D_E$ which are from the same source as
        a given target dataset (in case of \texttt{AMI}, \texttt{HASOC} and \texttt{HatEval}).
        }
\label{tab:setups}
\end{table}

In the following we describe our multi-dataset setup, i.e., the 9 corpora in
the external set ($D_E$) and the 13 target ($D_t$) datasets.
For a high-level overview of the setup, including labels of the external and
target datasets, we refer to Table~\ref{tab:setups}.
Note that although we list multiple target datasets, we build a dedicated model for
each of them separately ($M_t$) in step 2 using $M_E$ which is trained on all
the datasets in $D_E$ jointly in step 1.
Although
the goal of MDT is to allow for an easy external dataset selection
specific to a target task, we only consider a single external set for all target
datasets to save resources.
However, this general setup shows that an external set can be selected without
very careful data selection, and gives a lower bound of the achievable
performance as we expect more benefits from a more specialized setup.

The goal of the setup is to include a wide range of datasets related to
abusive language detection, such as hate speech,
offense, abuse,
sexism, racism detection as well as target identification.
Additionally, we include datasets from the same task category but with different
label sets, e.g., \texttt{HASOC \underline{fine-grained abusive}} (hate, offensive,
profane) and \texttt{SRW \underline{fine-grained abusive}} (sexism, racism, normal).
We only include English datasets in
$D_E$,
while we used both
English and non-English corpora (De, Hi, It, Pt-Br) as the target datasets to
test cross-lingual transfer as well.
Furthermore, we test on Stormfront which contains forum posts
instead of Twitter and Facebook messages as the datasets in $D_E$ do.
To avoid data leakage between the external train and the target test sets
(in case of \texttt{AMI}, \texttt{HASOC} and \texttt{HatEval}),
i.e., to filter samples which have the same input samples but with different labels or
inputs from different languages with the same labeling methodology, we remove
all datasets from $D_E$ which are from the same authors as the test set, e.g.,
we omit all \texttt{AMI} external datasets when training $M_E$ in step 1 in case we
test on \texttt{AMI binary misogyny It}.\footnote{Note that this filtering step requires
us to train separate $M_E$ models in these cases, resulting in
4
different $M_E$ models
overall.
}

\subsection{Compared Systems}
\label{sec:sec:baselines}

We compare
MDT to
four types of baseline
systems.
We use off-the-shelf pre-trained LMs and train them using the few-shot setup as
in the second step of our proposed approach without training them on the
external datasets (LM-base).
As shown by \citet{gururangan-etal-2020-dont} fine-tuning LMs on the domain of
the task of interest by further MLM training on unlabeled data can improve
down-stream task performance.
In order to test the effectiveness of this step in contrast to our approach
which leverages labels instead, we run MLM on the external datasets of the
above-mentioned setups for one epoch (MLM).
To test the importance of the two separate steps of our approach we
perform multitask learning, i.e., both the external datasets and the target
dataset are used in a single step similarly as in step 1 in
Section~\ref{sec:approach} (MTL).
Additionally, we test adapter-fusion \cite{pfeiffer-etal-2021-adapterfusion}
which eliminates negative task interference by first, training independent
adapters on
each dataset, followed by combining them for the target task (Fusion).\footnote{We use
prompt-training in contrast to the original work which uses classification heads.}

\paragraph{Model parameters}
We use \emph{xlm-roberta-base} as our base LM
\cite{conneau-etal-2020-unsupervised}
for all the baselines and our MDT setups as well.
In our early experiments we tested \emph{bert-base-multilingual-cased}
\cite{devlin-etal-2019-bert} as well, which resulted in similar conclusions.
However, XLM-R benefited slightly more from MDT which suggest that even larger
models might be able to exploit general information from external datasets to a
higher degree.
For evaluation we used macro averaged $F_1$ score averaged over 5 different
seeds\footnote{We use a single seed for MTL since it is trained on a large set
of inputs jointly.} in
order to reduce the high variance issue of few-shot classification
\cite{zheng-etal-2022-fewnlu}.
We follow the standard n-shot setting for few-shot learning.
Due to the high label bias of abusive language datasets, a large number of
samples have to be considered for annotation in order to increase the training
size of the minority classes with even a few examples.
Thus, we selected $n=4$ as a reasonable trade-off between costs and the
amount of training data for the target datasets (step 2).
We experiment with
different
$n$ values
shown in Figure~\ref{fig:nshot}.
We used the full training and validation sets of the external datasets (step 1).
For all datasets we use the official train, validation and test splits if given,
otherwise we take 80/20 train/test split of the full dataset and/or an
additional 80/20
split of the train set for final training and validation if the latter is not given.
For the implementation we used the
\emph{Huggingface transformers} \cite{wolf-etal-2020-transformers} and
\emph{OpenPrompt} \cite{ding-etal-2022-openprompt} libraries for
prompt-learning.
The used hyperparameters
are: batch size $1$,\footnote{Due to limited GPU memory, we could not test on larger batch sizes. } gradient accumulation steps $16$, warm-up steps $10$,
learning rate $5\times 10^{-5}$ and dropout $0.1$.
We ran
a single epoch
on each external dataset in step 1,
since we found that longer training made our models biased towards some of the
most frequent labels.
In contrast, we used early stopping on the
target $n$-shot validation set in step 2.

\begin{table*}[t]
\centering
    \begin{subtable}{0.89\linewidth}
        \resizebox{\textwidth}{!}{            \newcommand\myrow{
            \csvcoli & {\csvcolii}\stdstr{\csvcoliii} & {\csvcoliv}\stdstr{\csvcolv}
            & {\csvcolvi}\stdstr{\csvcolvii} & {\csvcolviii}\stdstr{\csvcolix} &
            {\csvcolx}\stdstr{\csvcolxi} & {\csvcolxii}\stdstr{\csvcolxiii}& {\csvcolxxix}
            }
        \begin{tabular}{r|lll|l|l|l|l}
        & \multicolumn{6}{c|}{fine-grained}&  \\
        \cline{2-7}
        & \multicolumn{3}{c|}{abusive}      & offensive & toxicity & target & \\
        & \multicolumn{3}{c|}{HASOC (1/3)}        & GermEval                      & ToLD-Br                      & OLID                        & \\
        & En                                & Hi                            & De                           & De (1/4)                          & Pt-Br (4/7)                     & En (1/3) & avg. \\
        \hline
        \hline
                \begin{filecontents*}{results_main.csv}
,hasoc en,std,hasoc hi,std,hasoc de,std,germeval fine,std,told-br,std,olid target,std,stormfront,std,hateval en,std,hateval es,std,olid binary,std,germeval binary,std,ami en,std,ami it,std,avg,avg fine,avg bin
LM-base,32.76,5.94,33.59,5.68,32.55,11.71,21.48,2.22,8.25,3.54,36.94,3.80,52.82,6.36,57.31,4.98,45.60,6.78,52.70,7.18,50.89,5.33,\textbf{57.31},4.25,60.33,10.45,41.73,27.59,53.85
MLM,35.98,7.70,\textbf{35.81},3.79,28.70,9.94,21.97,3.89,8.98,2.76,40.88,5.38,54.88,4.89,58.91,2.53,46.37,9.00,53.63,7.09,52.81,1.16,48.68,8.20,64.24,5.82,42.45,28.72,54.22
MTL,13.22,0.00,16.38,0.00,24.10,0.00,19.84,0.00,10.69,0.00,14.11,0.00,54.22,0.00,54.40,0.00,48.85,0.00,56.00,0.00,47.05,0.00,49.73,0.00,45.32,0.00,34.92,16.39,50.80
Fusion,37.35,8.06,34.29,4.14,29.18,8.09,21.33,3.88,8.26,1.20,42.45,4.03,46.79,7.50,53.91,4.41,45.94,1.57,64.67,2.70,56.95,1.95,34.95,5.30,39.11,11.17,39.63,28.81,48.91
MDT,\textbf{40.48},5.37,34.36,2.59,\textbf{33.96},7.83,\textbf{27.70},3.78,\textbf{12.83},1.78,\textbf{49.55},2.92,\textbf{60.41},5.75,\textbf{60.20},5.52,\textbf{54.47},0.79,\textbf{64.81},8.55,\textbf{65.02},4.17,47.77,1.36,\textbf{66.98},4.03,\textbf{47.58},\textbf{33.14},\textbf{59.95}
\end{filecontents*}
\csvreader[
                    late after line = \\
                ]{results_main.csv}{}{
                    \myrow
                }
        \end{tabular}        }
    \end{subtable}

    \vspace{5mm}
    \begin{subtable}{0.89\linewidth}
        \resizebox{\textwidth}{!}{            \newcommand\myrow{
            \csvcoli & {\csvcolxiv}\stdstr{\csvcolxv} & {\csvcolxvi}\stdstr{\csvcolxvii} &
            {\csvcolxviii}\stdstr{\csvcolxix} & {\csvcolxx}\stdstr{\csvcolxxi} &
            {\csvcolxxii}\stdstr{\csvcolxxiii} & {\csvcolxxiv}\stdstr{\csvcolxxv} &
            {\csvcolxxvi}\stdstr{\csvcolxxvii} & {\csvcolxxx}
            }
        \begin{tabular}{r|lll|ll|ll|l}
        & \multicolumn{7}{c|}{binary} &  \\
        \cline{2-8}
        &  \multicolumn{3}{c|}{hate} & \multicolumn{2}{c|}{offensive} & \multicolumn{2}{c|}{misogyny} & \\
        &  Stormfront                & \multicolumn{2}{c|}{HatEval}   & OLID                            & GermEval          & \multicolumn{2}{c|}{AMI} & \\
        &  En                              & En                & Es                         & En                & De                & En                & It & avg. \\
        \hline
        \hline
                \csvreader[
                    late after line = \\
                ]{results_main.csv}{}{
                    \myrow
                }
        \end{tabular}        }
    \end{subtable}
\caption{
    Macro averaged $F_1$ scores and standard deviation (\%) on the fine-grained
    and binary target datasets of our multi-dataset approach using 4-shot
    training.
            In case there are unseen labels in a given target dataset, we highlight them
    together with the overall number of labels in parentheses.
        The best result for each target dataset is in \textbf{bold}.
            }
\label{tab:main_res}
\end{table*}

\section{Results}
\label{sec:results}

First, we present our main results
followed by the analysis of
different few-shot sizes
and
the model performance on
each label separately.
Then we discuss an ablation study for
a better understanding of how the
external datasets affect the final performance.
Finally, we
briefly discuss our experiments on \textsc{HateCheck}.

Our main results with 4-shot training are presented in Table~\ref{tab:main_res}.
On a higher level it can be seen that
MDT improved over all baselines
in 11 out of 13 cases (in 12 cases compared to LM-base).
MDT is significantly better ($\alpha=0.05$) than LM-base on all datasets except
on \texttt{AMI En} and \texttt{HASOC De}.
We used the significance test
proposed in~\cite{dror-etal-2019-deep},
which accounts for the challenges of comparing deep neural networks, including the
difficulties due to the use of multiple random seeds.
The method compares the score distributions generated by different runs
(random seeds) of a given model type using an approach based on \emph{Almost
Stochastic Dominance} relation of the distributions.
The MLM, MTL and Fusion baselines also improve over the LM-base system,
however, not as consistently and to a lesser extent than our approach.
MTL and Fusion even achieve lower performance than LM-base when the averaged
performance over all datasets is considered.
This indicates that i) relying on the labels other than only the domain
adaptation effect of MLM is beneficial, ii) the two-step approach of MDT is
more effective, since the very low number of samples of the target dataset are
suppressed by the external data samples when they are added directly into MTL
and iii) the small number of target dataset samples is not enough to properly
fuse dataset specific adapters
(Fusion).
Additionally, training on the external datasets makes the models more consistent
over different runs, as shown by the decreased standard deviation values on
most of the datasets.

MDT achieves comparable average improvements on the fine-grained and the binary
target datasets.
Looking at the former set,
not only seen but unseen
labels as well were improved
(even on \texttt{ToLD-Br} with more than half of its
labels unseen), suggesting that the general abusive language aware
$M_E$ model helps learning the fine-grained label sets of these datasets even
with only a few-shots being available.
We discuss the improvements on the different labels in more details below in the
per label analysis section.
The only exception is the \texttt{HASOC fine-grained abusive Hi} dataset, where
the MLM baseline achieved the best results, although MDT also improved over
LM-base.
Our conjecture is that
this
is partly due to the high ratio of English content in
the dataset caused by its code-mixed nature, and as Table~\ref{tab:main_res}
suggests, MLM tends to perform better on English target datasets compared to
non-English sets.

\begin{table*}[t]
    \centering
    \newcommand\myrow{
    \csvcolvi & \perc{\csvcolvii} & \perc{\csvcolix}
    & \perc{\csvcolxi}
    }
    \begin{subtable}{0.39\linewidth}
        \resizebox{\textwidth}{!}{            \begin{tabular}{r|rrr}
                              & hate & offensive    & \textbf{profane} \\
                \hline
                \hline
                \begin{filecontents*}{results_hasoc_en-abusive.csv}
,,,,,,HATE_F1,HATE_F1_STD,OFFN_F1,OFFN_F1_STD,PRFN_F1,PRFN_F1_STD,macro_averaged_F1,macro_averaged_F1_STD
hasoc19_en-fine_grained,class,4-shot,bert,base,-,0.4644034188,0.08454248979,0.2168948975,0.1171043487,0.327050983,0.1554549616,0.3361164331,0.04217167599
hasoc19_en-fine_grained,class,4-shot,bert,fine2fine_cross1,-,0.4484171455,0.2303178228,0.2552981543,0.0753617229,0.6798661616,0.04625223986,0.4611938205,0.06912769616
hasoc19_en-fine_grained,class,4-shot,bert,fine2fine_cross1_MLM,-,0.4161001975,0.1358989044,0.08906157283,0.07988259618,0.3356106995,0.1719944171,0.2802574899,0.03922490971
hasoc19_en-fine_grained,class,4-shot,bert,fine2fine_cross1_JOINT,-,0.03125,0,0.2325581395,0,0.4451410658,0,0.2363164018,0
hasoc19_en-fine_grained,class,4-shot,bert,fine2fine_cross3,-,0.6721585603,0.09100609523,0.118081405,0.06445858086,0.6896309467,0.00957273585,0.493290304,0.0171891279
hasoc19_en-fine_grained,class,4-shot,bert,fine2fine_cross3_MLM,-,0.4312219331,0.1070444948,0.2028451133,0.0779907999,0.3601113676,0.09327404275,0.3313928047,0.01913514571
hasoc19_en-fine_grained,class,4-shot,bert,fine2fine_cross3_JOINT,-,0.4465116279,0,0.1851851852,0,0.3718592965,0,0.3345187032,0
hasoc19_en-fine_grained,class,4-shot,bert,fine2fine_cross3_ab1,-,0.4262605003,0.1956289679,0.2359711641,0.09827515292,0.5431029056,0.2526320181,0.40177819,0.1004832209
hasoc19_en-fine_grained,class,4-shot,bert,all2,-,0.6590326338,0.09824454361,0.1428576751,0.0820587614,0.6480666398,0.02228561009,0.4833189829,0.01595535552
hasoc19_en-fine_grained,class,4-shot,bert,all2_balanced,-,0.5440737261,0.1453511859,0.2077556571,0.1191147844,0.6449755017,0.1129135109,0.4656016283,0.0549430133
hasoc19_en-fine_grained,class,8-shot,bert,base,-,0.4608633826,0.1730050124,0.2746587469,0.1082093526,0.3929260934,0.1300500324,0.3761494077,0.05167156929
hasoc19_en-fine_grained,class,8-shot,bert,fine2fine_cross1,-,0.527496544,0.1798075367,0.2473955779,0.08702436778,0.5611425648,0.2439817322,0.4453448956,0.06754736425
hasoc19_en-fine_grained,class,8-shot,bert,fine2fine_cross3,-,0.633423925,0.1228896958,0.2105001697,0.1441284451,0.7097963754,0.01242111143,0.5179068234,0.05682956628
hasoc19_en-fine_grained,class,8-shot,bert,fine2fine_cross3_ab1,-,0.4403271401,0.1895055408,0.2045451638,0.1080585879,0.6897817416,0.01592071359,0.4448846818,0.03230154115
hasoc19_en-fine_grained,class,8-shot,bert,all2,-,0.6756334494,0.03641346261,0.1394245807,0.04601020389,0.6551380101,0.01822600118,0.4900653467,0.03034923931
hasoc19_en-fine_grained,P0,zero_shot,bert,untrained,-,0.124137931,0,0.4048338369,0,0.08,0,0.2029905893,0
hasoc19_en-fine_grained,P0,zero_shot,bert,fine2fine_cross1,-,0,0,0.3955431755,0,0,0,0.1318477252,0
hasoc19_en-fine_grained,P0,zero_shot,bert,fine2fine_cross3,-,0,0,0.3955431755,0,0,0,0.1318477252,0
hasoc19_en-fine_grained,P0,zero_shot,bert,fine2fine_cross3_ab1,-,0.106870229,0,0.4034090909,0,0,0,0.1700931066,0
hasoc19_en-fine_grained,P0,1-shot,bert,base,-,0.288386536,0.1449751016,0.3331082795,0.1033306672,0.2230789498,0.0838021208,0.2815245885,0.03675272818
hasoc19_en-fine_grained,P0,1-shot,bert,all2,-,0.6766819923,0.04761611834,0.1786466185,0.0506950236,0.5285248989,0.260443355,0.4612845032,0.08469366807
hasoc19_en-fine_grained,P0,1-shot,xlmr,base,LM-base,0.2037398377,0.1645702992,0.3425716893,0.02386429712,0.2590469988,0.1228449538,0.2684528419,0.03803977885
hasoc19_en-fine_grained,P0,1-shot,xlmr,all2,MDT,0.3179644589,0.2111733422,0.3037406809,0.1009514627,0.3440946403,0.2076787505,0.32193326,0.02873701393
hasoc19_en-fine_grained,P0,4-shot,bert,base,-,0.2909763477,0.1210897687,0.3729141876,0.01764776548,0.2887977883,0.1258122349,0.3175627745,0.02016857054
hasoc19_en-fine_grained,P0,4-shot,bert,fine2fine_cross1,-,0.6246291375,0.08904132694,0.2193534516,0.08703073347,0.5097643026,0.2311842431,0.4512489639,0.06096801074
hasoc19_en-fine_grained,P0,4-shot,bert,fine2fine_cross1_MLM,-,0.4094634305,0.0883664673,0.2790513917,0.1290251067,0.3149057662,0.1065386998,0.3344735295,0.04715227577
hasoc19_en-fine_grained,P0,4-shot,bert,fine2fine_cross1_JOINT,-,0.01587301587,0,0.3977591036,0,0,0,0.1378773732,0
hasoc19_en-fine_grained,P0,4-shot,bert,fine2fine_cross3,-,0.6892445178,0.04666557625,0.1495905823,0.03725538158,0.5460680336,0.1942576857,0.4616343779,0.07131513304
hasoc19_en-fine_grained,P0,4-shot,bert,fine2fine_cross3_MLM,-,0.3295437158,0.1330659594,0.2185665376,0.1151668471,0.3227652556,0.1351521756,0.2902918363,0.01794581568
hasoc19_en-fine_grained,P0,4-shot,bert,fine2fine_cross3_JOINT,-,0,0,0.3955431755,0,0,0,0.1318477252,0
hasoc19_en-fine_grained,P0,4-shot,bert,fine2fine_cross3_ab1,-,0.6804321709,0.01880873052,0.1263002726,0.07970077829,0.6525237739,0.04699592232,0.4864187391,0.0418050574
hasoc19_en-fine_grained,P0,4-shot,bert,all2,-,0.7115811973,0.01874746456,0.1096530728,0.04100711535,0.6849374202,0.0244551976,0.5020572301,0.01651534007
hasoc19_en-fine_grained,P0,4-shot,bert,all2_balanced,-,0.6449850312,0.1630681744,0.2535823332,0.09844690658,0.5832908238,0.1378409791,0.4939527294,0.03581465583
hasoc19_en-fine_grained,P0,4-shot,bert,all2_glob_balanced_mlm_1_0,-,0.6368460822,0.03041238559,0.1747527769,0.07160204739,0.4972031379,0.1332778075,0.4362673323,0.03320092701
hasoc19_en-fine_grained,P0,4-shot,bert,all2_ab1,-,0.6994358801,0.04100468691,0.08605673365,0.06301148075,0.6246659486,0.1382985699,0.4700528541,0.0601022658
hasoc19_en-fine_grained,P0,4-shot,bert,all2_ab1_balanced,-,0.6866580145,0.02398120899,0.2297015802,0.06510193378,0.6672784212,0.05428435936,0.5278793386,0.03142712977
hasoc19_en-fine_grained,P0,4-shot,bert,all2_MLM,-,0.1354538649,0.09523969653,0.1690847062,0.122212181,0.4391873442,0.09427775159,0.2479086384,0.02969929691
hasoc19_en-fine_grained,P0,4-shot,bert,all2_JOINT,-,0,0,0.3955431755,0,0,0,0.1318477252,0
hasoc19_en-fine_grained,P0,4-shot,bert,all2_cross,-,0.6656563793,0.03288649064,0.1601380074,0.08588489317,0.5063768446,0.2544515927,0.4440570771,0.06279746989
hasoc19_en-fine_grained,P0,4-shot,xlmr,base,LM-base,0.3064231321,0.1756628719,0.2438503235,0.0898739243,0.432415936,0.0510757608,0.3275631305,0.05941763046
hasoc19_en-fine_grained,P0,4-shot,xlmr,all2,MLM,0.3834822675,0.2108391771,0.3190989179,0.05888572749,0.3767068885,0.1162689278,0.3597626913,0.0770116486
hasoc19_en-fine_grained,P0,4-shot,xlmr,all2,MTL,0,0,0.3966480447,0,0,0,0.1322160149,0
hasoc19_en-fine_grained,P0,4-shot,xlmr,all2,Fusion,0.5072185026,0.1325341277,0.1813622394,0.1169993345,0.4319909949,0.1641556261,0.3735239123,0.08056704869
hasoc19_en-fine_grained,P0,4-shot,xlmr,all2,MDT,0.5293705891,0.06677094121,0.2663920494,0.1334360872,0.4185440551,0.1739206649,0.4047688978,0.05374027212
hasoc19_en-fine_grained,P0,4-shot,xlmr,all2,MDT-abl,0.4993625991,0.1025841753,0.1901148083,0.05600866074,0.4192963323,0.1371524129,0.3695912466,0.02766716452
hasoc19_en-fine_grained,P0,4-shot,xlmr,all2_ab1,-,0.7108209278,0.0241855999,0.1424866132,0.1199071445,0.608560681,0.2369217795,0.4872894074,0.06207457698
hasoc19_en-fine_grained,P0,8-shot,bert,base,-,0.3916821138,0.2278604545,0.0877124348,0.05310049805,0.4103756741,0.1611632342,0.2965900743,0.07344538878
hasoc19_en-fine_grained,P0,8-shot,bert,fine2fine_cross1,-,0.6840407952,0.04489105251,0.08666589975,0.06461994745,0.6628574028,0.03002010254,0.4778546992,0.02956869778
hasoc19_en-fine_grained,P0,8-shot,bert,fine2fine_cross3,-,0.693920404,0.03405199314,0.1388558016,0.06788766067,0.6611063974,0.08864038454,0.4979608677,0.03614615944
hasoc19_en-fine_grained,P0,8-shot,bert,fine2fine_cross3_ab1,-,0.6214645979,0.117965392,0.2210096986,0.08174853491,0.67367508,0.05612534694,0.5053831255,0.03947858009
hasoc19_en-fine_grained,P0,8-shot,bert,all2,-,0.6790451055,0.01740597023,0.1247094602,0.0352340443,0.6989799122,0.0193787102,0.5009114927,0.0146028958
hasoc19_en-fine_grained,P0,8-shot,bert,all2_ab1,-,0.7051406502,0.04748905809,0.1123738361,0.05432955669,0.6107081622,0.1378535655,0.4760742161,0.04577916767
hasoc19_en-fine_grained,P0,8-shot,bert,all2_MLM,-,0.4645506708,0.1606198857,0.2586868707,0.0597230939,0.4521558041,0.143542844,0.3917977818,0.05624384934
hasoc19_en-fine_grained,P0,8-shot,bert,all2_JOINT,-,0,0,0.3955431755,0,0,0,0.1318477252,0
hasoc19_en-fine_grained,P0,8-shot,xlmr,base,LM-base,0.3609094952,0.1850443872,0.2811924366,0.08967774711,0.4968458464,0.05982170603,0.3796492594,0.05322062031
hasoc19_en-fine_grained,P0,8-shot,xlmr,all2,MDT,0.5444073034,0.08981883624,0.2354989562,0.1351512883,0.4243701073,0.08612892604,0.4014254556,0.03477147937
hasoc19_en-fine_grained,P0,16-shot,bert,base,-,0.4900622265,0.1170950268,0.261593144,0.08916675298,0.4308075121,0.1465432363,0.3941542942,0.06010706023
hasoc19_en-fine_grained,P0,16-shot,bert,all2,-,0.709434256,0.04657279614,0.1319056003,0.1063801135,0.7187178957,0.01053805104,0.5200192507,0.01873281756
hasoc19_en-fine_grained,P0,16-shot,xlmr,base,LM-base,0.5237017696,0.1233750188,0.2665960583,0.0591173436,0.5461815136,0.07426630849,0.4454931138,0.03797260826
hasoc19_en-fine_grained,P0,16-shot,xlmr,all2,MDT,0.6235304149,0.05447903702,0.2320624909,0.05763177778,0.5636669086,0.08981092842,0.4730866048,0.02234930225
hasoc19_en-fine_grained,P0,32-shot,bert,base,-,0.4747916441,0.1744119421,0.2192942244,0.09753837647,0.4685313058,0.1064960257,0.3875390581,0.05373682433
hasoc19_en-fine_grained,P0,32-shot,bert,all2,-,0.7424980234,0.011459031,0.09468782539,0.04779132819,0.7218941793,0.01657001191,0.5196933427,0.01609096597
hasoc19_en-fine_grained,P0,32-shot,xlmr,base,LM-base,0.6008047896,0.1124836987,0.1843007942,0.1246376943,0.5476775034,0.0913438831,0.444261029,0.0321502566
hasoc19_en-fine_grained,P0,32-shot,xlmr,all2,MDT,0.6153159601,0.08389806267,0.1826470172,0.149757845,0.5528458447,0.06106809884,0.4502696073,0.02200489387
hasoc19_en-fine_grained,P0,64-shot,bert,base,-,0.5680063361,0.07584863718,0.3137196964,0.05983939913,0.4363969881,0.1656216717,0.4393743402,0.06630949742
hasoc19_en-fine_grained,P0,64-shot,bert,all2,-,0.7153210292,0.04069358101,0.1092599957,0.09224785678,0.59494809,0.1892288043,0.4731763716,0.07979744524
hasoc19_en-fine_grained,P0,64-shot,xlmr,base,LM-base,0.3985896141,0.2303223215,0.3031190527,0.05617013856,0.5382070483,0.09362911538,0.4133052384,0.1085592622
hasoc19_en-fine_grained,P0,64-shot,xlmr,all2,MDT,0.6006766411,0.04165366789,0.2558562758,0.07611690047,0.4123124376,0.2392308058,0.4229484515,0.1014004487
\end{filecontents*}
\csvreader[range = {40,44}, late after line = \\]{results_hasoc_en-abusive.csv}{}{
                    \myrow
                }
            \end{tabular}        }
        \caption{HASOC fine-grained abusive En}
            \end{subtable}
        \hspace{5mm}
    \renewcommand\myrow{
    \csvcolvi & \perc{\csvcolvii} &
    \perc{\csvcolix} & \perc{\csvcolxi}
    }
    \begin{subtable}{0.37\linewidth}
        \resizebox{\textwidth}{!}{            \begin{tabular}{r|rrr}
                        & group & individual    & \textbf{other} \\
                \hline
                \hline
                \begin{filecontents*}{results_olid_fine_grained_target.csv}
,,,,,,GRP_F1,GRP_F1_STD,IND_F1,IND_F1_STD,OTH_F1,OTH_F1_STD,macro_averaged_F1,macro_averaged_F1_STD
OLID_target,class,4-shot,bert,base,-,0.3097667438,0.1543641178,0.4656110928,0.08980962823,0.2384941995,0.05045056081,0.3379573454,0.03835154788
OLID_target,class,4-shot,bert,target1,-,0.608597777,0.05480380307,0.6692007141,0.07336273699,0.1484601945,0.06345825228,0.4754195618,0.03717684502
OLID_target,class,4-shot,bert,target1_MLM,-,0.3429355876,0.1127831748,0.4335771711,0.2011804619,0.1005325207,0.06303965565,0.2923484265,0.03286908081
OLID_target,class,4-shot,bert,target1_JOINT,-,0,0,0.6389776358,0,0,0,0.2129925453,0
OLID_target,class,4-shot,bert,target1_ab1,-,0.4502156242,0.1601301449,0.6222665417,0.1142675429,0.2240697313,0.07536891927,0.4321839657,0.04382118389
OLID_target,class,4-shot,bert,all1,-,0.5858817478,0.03581913627,0.5247776434,0.04787516054,0.2092721997,0.09819844865,0.439977197,0.03158842343
OLID_target,class,4-shot,bert,all1_MLM,-,0.29412341,0.1463699045,0.4085595864,0.1465009145,0.1980306763,0.02758619186,0.3002378909,0.04155094139
OLID_target,class,4-shot,bert,all1_JOINT,-,0.2903225806,0,0.6142322097,0,0,0,0.3015182635,0
OLID_target,class,4-shot,bert,all2,-,0.5418035633,0.03826755522,0.4828813718,0.08272243664,0.1789853953,0.07525830361,0.4012234435,0.05233850609
OLID_target,class,4-shot,bert,all2_balanced,-,0.4413550184,0.08737742025,0.4428658384,0.2123885486,0.2496508887,0.067047219,0.3779572485,0.07262425877
OLID_target,class,8-shot,bert,base,-,0.3372148825,0.07729431838,0.5069852257,0.07561563069,0.2198059784,0.05558396533,0.3546686955,0.03037846667
OLID_target,class,8-shot,bert,target1,-,0.5488147496,0.1069124077,0.6959396277,0.04970175243,0.1579729883,0.134158561,0.4675757885,0.02148721273
OLID_target,class,8-shot,bert,target1_ab1,-,0.500930367,0.2282702367,0.6078498481,0.1334036414,0.1529312256,0.127366846,0.4205704802,0.06257061945
OLID_target,class,8-shot,bert,all1,-,0.5954479861,0.02325022112,0.547884416,0.04366283572,0.1818448245,0.08195629353,0.4417257422,0.03149261177
OLID_target,class,8-shot,bert,all2,-,0.5134487134,0.1028954163,0.4564098853,0.12101138,0.2358077667,0.02781100249,0.4018887884,0.07246391752
OLID_target,P0,zero_shot,bert,untrained,-,0.2916666667,0,0,0,0.2527472527,0,0.1814713065,0
OLID_target,P0,zero_shot,bert,target1,-,0.6467661692,0,0.630952381,0,0.1754385965,0,0.4843857155,0
OLID_target,P0,zero_shot,bert,target1_ab1,-,0.6025104603,0,0.4827586207,0,0,0,0.3617563603,0
OLID_target,P0,zero_shot,bert,target1_ab2,-,0.6291079812,0,0.6545454545,0,0.08333333333,0,0.4556622564,0
OLID_target,P0,zero_shot,bert,target1_ab3,-,0.02409638554,0,0.231884058,0,0.1470588235,0,0.1343464223,0
OLID_target,P0,zero_shot,bert,all1,-,0.4869565217,0,0.1308411215,0,0.2941176471,0,0.3039717634,0
OLID_target,P0,1-shot,bert,base,-,0.4093002876,0.09219773749,0.5859639449,0.09249255031,0.1631045772,0.1026715905,0.3861229365,0.05520855833
OLID_target,P0,1-shot,bert,all2,-,0.6040101102,0.05099490644,0.5948886748,0.09535511173,0.1637019153,0.0951753804,0.4542002334,0.04717630877
OLID_target,P0,1-shot,xlmr,base,LM-base,0.4695542099,0.09063780441,0.4339430521,0.08115731446,0.07899245852,0.09760303873,0.3274965735,0.03479211859
OLID_target,P0,1-shot,xlmr,all2,MDT,0.6258452879,0.03501528367,0.5741531002,0.1312487696,0.2002355367,0.07917763256,0.4667446416,0.0444725612
OLID_target,P0,4-shot,bert,base,-,0.4298098757,0.09611688014,0.5188895327,0.1152349737,0.2032107014,0.06439396579,0.3839700366,0.05304591871
OLID_target,P0,4-shot,bert,target1,-,0.5839786513,0.05642079124,0.6318086129,0.05083010007,0.2078380036,0.03743395834,0.474541756,0.02171598588
OLID_target,P0,4-shot,bert,target1_MLM,-,0.3893951107,0.07899610159,0.5381173384,0.04941398862,0.2041039835,0.04895442698,0.3772054775,0.03032991321
OLID_target,P0,4-shot,bert,target1_JOINT,-,0.6341463415,0,0.6774193548,0,0,0,0.4371885654,0
OLID_target,P0,4-shot,bert,target1_ab1,-,0.5911289692,0.05027955391,0.5549598948,0.1199800198,0.2273171526,0.05415426694,0.4578020055,0.04344528035
OLID_target,P0,4-shot,bert,target1_ab2,-,0.6202508197,0.02998455893,0.6006973274,0.1286178848,0.2172771985,0.0449882555,0.4794084485,0.03897721921
OLID_target,P0,4-shot,bert,target1_ab3,-,0.3988702863,0.2323963394,0.3053274239,0.20260945,0.1409424856,0.04199598509,0.2817133986,0.1334451073
OLID_target,P0,4-shot,bert,all1,-,0.6284413934,0.02558344831,0.6108939732,0.06199504898,0.2260177163,0.1100837114,0.4884510277,0.02904350416
OLID_target,P0,4-shot,bert,all1_MLM,-,0.3280327453,0.1156042021,0.5485569275,0.09855514786,0.1275856195,0.09174714419,0.3347250974,0.03606378892
OLID_target,P0,4-shot,bert,all1_JOINT,-,0,0,0,0,0.2822580645,0,0.09408602151,0
OLID_target,P0,4-shot,bert,all2,-,0.5723790405,0.08515980521,0.648560255,0.0531250144,0.1867444607,0.1006159478,0.4692279187,0.02759431548
OLID_target,P0,4-shot,bert,all2_balanced,-,0.4902272232,0.1001686501,0.5817084478,0.06642897792,0.1630530869,0.0631834505,0.4116629193,0.03358597412
OLID_target,P0,4-shot,bert,all2_glob_balanced_mlm_1_0,-,0.5954364081,0.04535003966,0.6003586926,0.09590602906,0.2465102601,0.04951737667,0.4807684536,0.04914164603
OLID_target,P0,4-shot,bert,all2_ab1,-,0.6214234683,0.02608672859,0.6660247158,0.05709628111,0.2264687437,0.06907135395,0.5046389759,0.02169174042
OLID_target,P0,4-shot,bert,all2_ab1_balanced,-,0.40052583,0.1141778027,0.5827413618,0.1362847571,0.175130928,0.06558847956,0.3861327066,0.04566990278
OLID_target,P0,4-shot,bert,all2_MLM,-,0.3376233127,0.1225355009,0.3946211859,0.1719855564,0.2463732679,0.01728261451,0.3262059222,0.07221711346
OLID_target,P0,4-shot,bert,all2_JOINT,-,0.4622641509,0,0.4155844156,0,0.1,0,0.3259495222,0
OLID_target,P0,4-shot,xlmr,base,LM-base,0.418451964,0.0931231803,0.5105783782,0.08899049376,0.1791725152,0.07841072342,0.3694009525,0.03798577537
OLID_target,P0,4-shot,xlmr,all2,MLM,0.4716999261,0.06128386908,0.5748207561,0.06377632591,0.1798986211,0.07069982396,0.4088064345,0.05378770968
OLID_target,P0,4-shot,xlmr,all2,MTL,0.1411764706,0,0,0,0.2821576763,0,0.1411113823,0
OLID_target,P0,4-shot,xlmr,all2,Fusion,0.5821251185,0.0319106159,0.5147871792,0.1050049685,0.1765519199,0.08318261879,0.4244880725,0.04032647764
OLID_target,P0,4-shot,xlmr,all2,MDT,0.6007054581,0.1150304771,0.6568735877,0.08592461939,0.2289038019,0.06604243477,0.4954942826,0.02923825792
OLID_target,P0,4-shot,xlmr,all2,MDT-abl,0.6079773692,0.105428996,0.5706645979,0.146649159,0.1632681869,0.06932318197,0.4473033847,0.04154941286
OLID_target,P0,4-shot,xlmr,all2_ab1,-,0.6547321558,0.04002665307,0.7245660383,0.03226260178,0.2242823869,0.1129828242,0.5345268604,0.03394720559
OLID_target,P0,8-shot,bert,base,-,0.4133467289,0.1083582591,0.5097686287,0.2187174844,0.140390799,0.06812727485,0.3545020522,0.08358390887
OLID_target,P0,8-shot,bert,target1,-,0.558308909,0.08004281737,0.6475966795,0.02603819783,0.2443639063,0.02939555161,0.4834231649,0.01749024904
OLID_target,P0,8-shot,bert,target1_ab1,-,0.5819050828,0.09192477173,0.6732028722,0.05913842102,0.2945906663,0.06165658087,0.5165662071,0.02059797009
OLID_target,P0,8-shot,bert,target1_ab2,-,0.6078961557,0.05151541423,0.6254475623,0.05790410373,0.2301843318,0.06835256506,0.4878426833,0.0252317995
OLID_target,P0,8-shot,bert,target1_ab3,-,0.3170519777,0.2494187716,0.4772258587,0.1294044899,0.1653087933,0.07597182523,0.3198622099,0.1237849325
OLID_target,P0,8-shot,bert,all1,-,0.6536918383,0.02968739882,0.6522141527,0.05441461817,0.2506431867,0.08528794383,0.5188497259,0.02003564986
OLID_target,P0,8-shot,bert,all2,-,0.6394816419,0.03442355843,0.6739797855,0.04356867518,0.1673853857,0.1174375252,0.4936156044,0.03189311022
OLID_target,P0,8-shot,bert,all2_ab1,-,0.614365414,0.05948933523,0.6780641893,0.0503508843,0.2448342067,0.1023319487,0.51242127,0.02458464688
OLID_target,P0,8-shot,bert,all2_MLM,-,0.2869588033,0.1265757597,0.5482596885,0.06426737551,0.2003470772,0.0545098076,0.345188523,0.04451358983
OLID_target,P0,8-shot,bert,all2_JOINT,-,0.6048780488,0,0.6559139785,0,0,0,0.4202640091,0
OLID_target,P0,8-shot,xlmr,base,LM-base,0.4917162109,0.05581594947,0.4901507156,0.1579030608,0.2161456321,0.06070621239,0.3993375195,0.04800668976
OLID_target,P0,8-shot,xlmr,all2,MDT,0.6700288424,0.03347131522,0.6597961383,0.08364370786,0.2150043411,0.07273253821,0.5149431073,0.03658137059
\end{filecontents*}
\csvreader[range = {43, 47}, late after line = \\]{results_olid_fine_grained_target.csv}{}{
                    \myrow
                }

            \end{tabular}        }
        \caption{OLID fine-grained target En}
            \end{subtable}
    \vspace{0.5cm}

    \renewcommand\myrow{
    \csvcolvi & \perc{\csvcolxiii} &
    \perc{\csvcolxix} & \perc{\csvcolix} &
    \perc{\csvcolxxi} & \perc{\csvcolvii} &
    \perc{\csvcolxvii} & \perc{\csvcolxv}
    }
    \begin{subtable}{0.80\linewidth}
        \resizebox{\columnwidth}{!}{            \begin{tabular}{r|rrrrrrr}
            & misogyny & racism & \textbf{insult} & \textbf{xenophobia} &
                       \textbf{LGBTQ+phobia} & \textbf{obscene} & normal \\
                \hline
                \hline
                \begin{filecontents*}{results_told-br_fine_grained.csv}
,,,,,,homophobia_F1,homophobia_F1_STD,insult_F1,insult_F1_STD,macro_averaged_F1,macro_averaged_F1_STD,misogyny_F1,misogyny_F1_STD,none_F1,none_F1_STD,obscene_F1,obscene_F1_STD,racism_F1,racism_F1_STD,xenophobia_F1,xenophobia_F1_STD
told_br_fine_grained,class,4-shot,bert,base,-,0.009313243941,0.009744127208,0.05389181223,0.03976005089,0.08251596065,0.02112734093,0.001575397814,0.001971222713,0.4220211102,0.2079757188,0.08985207653,0.07159751937,0,0,0.0009580838323,0.001916167665
told_br_fine_grained,class,4-shot,bert,sex_fine_grained1,-,0.007193773968,0.009229727266,0.08113645715,0.04367132538,0.1017692454,0.03204426128,0,0,0.5023926558,0.2455620538,0.1204689852,0.03359992794,0.001192845427,0.001786646761,0,0
told_br_fine_grained,class,4-shot,bert,sex_fine_grained1_balanced,-,0.01624378097,0.004770600454,0.04566370851,0.03461946531,0.06840287355,0.04143883676,0.0008456659619,0.001691331924,0.2859973388,0.297044607,0.1263202635,0.02531110107,0.002136453857,0.001990084536,0.001612903226,0.003225806452
told_br_fine_grained,class,4-shot,bert,sex_fine_grained1_MLM,-,0.0146787918,0.01350437361,0.05890548878,0.04640590096,0.08719901724,0.03183619493,0.001705372002,0.002277200796,0.4075923376,0.2283250118,0.1273228952,0.04260407784,0.0001882352941,0.0003764705882,0,0
told_br_fine_grained,class,4-shot,bert,sex_fine_grained1_JOINT,-,0,0,0,0,0.1308284903,0,0,0,0.9157994324,0,0,0,0,0,0,0
told_br_fine_grained,class,4-shot,bert,all2,-,0.03162648213,0.03604770729,0.1242374227,0.01822896099,0.07957153509,0.02051451005,0,0,0.284450441,0.1696512046,0.1128511059,0.0333408997,0,0,0.003835293922,0.003814784605
told_br_fine_grained,class,4-shot,bert,all2_balanced,-,0.01250964427,0.004630382476,0.08194487087,0.01743271085,0.06636319794,0.02536960208,0.001233704369,0.001917334512,0.2543494305,0.1943336772,0.1080979063,0.02473170314,0.002583229242,0.003350159978,0.003823600054,0.002085078553
told_br_fine_grained,class,8-shot,bert,base,-,0.004641158777,0.005691207916,0.07836377748,0.0418769976,0.08250710472,0.02457398195,0.0008130081301,0.00162601626,0.4285525275,0.2210434589,0.06467642454,0.07484299313,0.0005028366136,0.0007031755693,0,0
told_br_fine_grained,class,8-shot,bert,sex_fine_grained1,-,0.003980099502,0.007960199005,0.08898321345,0.05669335503,0.1038537091,0.03632589506,0.001459854015,0.002919708029,0.5441953825,0.2310443423,0.08624529986,0.0493970293,0.001732006928,0.002366298036,0.0003801076972,0.0007602153944
told_br_fine_grained,class,8-shot,bert,all2,-,0.0149330592,0.007881745141,0.1254146054,0.04113860994,0.08215595244,0.019437634,0.00131147541,0.00262295082,0.3156697396,0.1386975636,0.1108988495,0.03172786456,0.001245102597,0.001449954809,0.0056188353,0.006765225583
told_br_fine_grained,P0,zero_shot,bert,untrained,-,0,0,0.05517241379,0,0.01085839273,0,0.003009027081,0,0.004122497055,0,0.01033591731,0,0.00336889388,0,0,0
told_br_fine_grained,P0,zero_shot,bert,sex_fine_grained1,-,0,0,0,0,0.1385537821,0,0.0618556701,0,0.9080208048,0,0,0,0,0,0,0
told_br_fine_grained,P0,1-shot,bert,base,-,0.01469540281,0.005691026752,0.04400241127,0.0242787212,0.07560299331,0.03491177672,0.006990647834,0.008674010385,0.3811295441,0.2519823024,0.08149933269,0.01180025841,0,0,0.0009036144578,0.001807228916
told_br_fine_grained,P0,1-shot,bert,all2,-,0.01424473831,0.02394252736,0.0387422115,0.03919144099,0.1106989739,0.02490923906,0.01061265156,0.00385075329,0.6156649299,0.2182560488,0.09316776735,0.04395571692,0,0,0.00246051862,0.002026689013
told_br_fine_grained,P0,1-shot,xlmr,base,LM-base,0.01595039298,0.01510285663,0.09335577186,0.02718185838,0.08690125677,0.02584681381,0.009039322026,0.009505600672,0.3846087103,0.1140097095,0.09863668229,0.05414062114,0.005840724947,0.002869432287,0.0008771929825,0.001754385965
told_br_fine_grained,P0,1-shot,xlmr,all2,MDT,0.0183019462,0.01774945716,0.07976072024,0.07224888823,0.1109015429,0.01420932535,0.01400159094,0.01036140987,0.4335432694,0.1139120047,0.2256922262,0.04415088226,0.001086956522,0.002173913043,0.003924091064,0.004712087259
told_br_fine_grained,P0,4-shot,bert,base,-,0.01005320811,0.009026061826,0.06351485171,0.04209550422,0.1011892023,0.01576908994,0.02191843244,0.04170607698,0.5403516034,0.1289458856,0.07097024904,0.03857480469,0.001516071112,0.00203374071,0,0
told_br_fine_grained,P0,4-shot,bert,sex_fine_grained1,-,0,0,0,0,0.1367242477,0.0004481630862,0.05114874253,0.002580880599,0.9059209915,0.0005745853579,0,0,0,0,0,0
told_br_fine_grained,P0,4-shot,bert,sex_fine_grained1_balanced,-,0.01659131425,0.01061034023,0.1709935488,0.01522812618,0.09502657953,0.02762739381,0.009742606304,0.002742220988,0.3075875757,0.1830972874,0.1497543187,0.01440723497,0.006994469653,0.006225953304,0.00352222326,0.002548320323
told_br_fine_grained,P0,4-shot,bert,sex_fine_grained1_mlm-0_5,-,0.01637505263,0.01265023571,0.07012234179,0.04540519374,0.07636086862,0.02700134213,0.01373518855,0.01170306932,0.3312505865,0.2685604605,0.09523492714,0.05407557076,0.001184495472,0.00145089616,0.006623488208,0.006263217812
told_br_fine_grained,P0,4-shot,bert,sex_fine_grained1_mlm-1_0,-,0.02025737586,0.01688607295,0.09043844896,0.04986474445,0.09286449336,0.03891998646,0.0176763911,0.01005213968,0.4136555185,0.2981773576,0.07172843349,0.04468688358,0.01152569634,0.01418230194,0.02476958928,0.02421869947
told_br_fine_grained,P0,4-shot,bert,sex_fine_grained1_MLM,-,0.01356397395,0.01098900729,0.05284462752,0.03179734827,0.1161708323,0.01834946862,0,0,0.6599360184,0.126195385,0.0868512062,0.03515827578,0,0,0,0
told_br_fine_grained,P0,4-shot,bert,sex_fine_grained1_JOINT,-,0,0,0,0,0.12335312,0,0,0,0.8634718402,0,0,0,0,0,0,0
told_br_fine_grained,P0,4-shot,bert,all2,-,0.006108538666,0.005082068577,0.04488141505,0.03177160748,0.09549812433,0.02184848059,0.01300537199,0.006808728091,0.4874683755,0.2213137767,0.1037225887,0.04877617111,0,0,0.0133005804,0.01967563941
told_br_fine_grained,P0,4-shot,bert,all2_ne-1,-,0.01687432538,0.003287770221,0.07660979964,0.02336067981,0.078329809,0.01129526733,0.01339963978,0.00825841295,0.3015791844,0.08200831865,0.1290267958,0.0375273491,0.003663711388,0.004116941413,0.007155206589,0.004104692651
told_br_fine_grained,P0,4-shot,bert,all2_ne-1_glob_balanced_mlm_1_0,-,0.01041802651,0.009935254618,0.08852065372,0.06320918181,0.07160910602,0.01786970838,0.01192929226,0.0105710149,0.3006189851,0.1023414849,0.07750917759,0.04464312196,0.006158698575,0.006664163791,0.00610890838,0.007989728927
told_br_fine_grained,P0,4-shot,bert,all2_balanced,-,0.01680421033,0.01190952865,0.05018451052,0.03559340439,0.0920055179,0.02657787102,0.01236324649,0.007977145597,0.43907141,0.1287493193,0.1202624773,0.062353935,0.001330033251,0.001934563807,0.004022737434,0.003749112672
told_br_fine_grained,P0,4-shot,bert,all2_ab1,-,0.0121037997,0.01575538506,0.02400649289,0.03189491037,0.09768041973,0.02246351292,0.06579737459,0.08105888344,0.4191870406,0.1045430641,0.158440353,0.01361492289,0.0005263157895,0.001052631579,0.003701561596,0.004664394483
told_br_fine_grained,P0,4-shot,bert,all2_ab1_balanced,-,0.00451735953,0.004268711014,0.08391723638,0.04858592782,0.09450366146,0.01683703194,0.06597929949,0.09365431022,0.4184824466,0.1819141115,0.08420620155,0.06753105379,0.003082209344,0.005267173959,0.001340877351,0.001953563163
told_br_fine_grained,P0,4-shot,bert,all2_mlm_1_0,-,0.01349405067,0.007959647431,0.05894437393,0.04032832324,0.07927297088,0.03428552652,0.0192571822,0.00963404689,0.3717508467,0.3180908667,0.07621034798,0.05619778804,0.001000172719,0.0009103208502,0.0142538219,0.01878819455
told_br_fine_grained,P0,4-shot,bert,all2_glob_balanced_mlm_1_0,-,0.01240626103,0.008710955219,0.09556882511,0.02553949358,0.1173151121,0.01381560236,0.001623581665,0.001560943185,0.520855427,0.1247880198,0.1568439074,0.0591742136,0.003937920234,0.002216867313,0.02996986253,0.03433127927
told_br_fine_grained,P0,4-shot,bert,all2_balanced_mlm_1_0,-,0.009717404483,0.009674668414,0.04031029104,0.02097006833,0.05515451327,0.04369787934,0.004848123883,0.006809538279,0.2654880436,0.3317636629,0.05703995017,0.05330894795,0.002974165568,0.001915767931,0.005703614215,0.006905980313
told_br_fine_grained,P0,4-shot,bert,all2_MLM,-,0.02301751341,0.0132324739,0.08440941911,0.01491697998,0.1003084181,0.02938550352,0.003714727881,0.003476611784,0.4777979451,0.2290360955,0.110782315,0.02244885728,0.0008109895894,0.001014717897,0.00162601626,0.00325203252
told_br_fine_grained,P0,4-shot,bert,all2_JOINT,-,0.06779661017,0,0.01606425703,0,0.1422991177,0,0,0,0.9122329569,0,0,0,0,0,0,0
told_br_fine_grained,P0,4-shot,xlmr,base,LM-base,0.02322011966,0.01246001635,0.09596150417,0.02678848203,0.08248222734,0.03536098287,0.02842732795,0.0414361395,0.2765384702,0.1787000954,0.1417282238,0.02227662593,0.003585757031,0.001991633724,0.007914188553,0.01074271093
told_br_fine_grained,P0,4-shot,xlmr,all2,MLM,0.0275926708,0.0203651659,0.08756258377,0.0501930202,0.08984154452,0.02764951359,0.0130617425,0.009601284949,0.3383313972,0.1674918441,0.1500761806,0.0273607647,0.00877391613,0.004212259954,0.003492320659,0.004754050058
told_br_fine_grained,P0,4-shot,xlmr,all2,MTL,0.02247191011,0,0,0,0.1068816764,0,0,0,0.5679855393,0,0.1577142857,0,0,0,0,0
told_br_fine_grained,P0,4-shot,xlmr,all2,Fusion,0.02733631875,0.008166565587,0.08712359278,0.02590263173,0.08256855271,0.01197620559,0.007541329095,0.004082465347,0.3747745245,0.02099534929,0.07742653684,0.02997252043,0.003103030757,0.001001859814,0.0006745362563,0.001349072513
told_br_fine_grained,P0,4-shot,xlmr,all2,MDT,0.0371753366,0.03621199192,0.1247845024,0.06333169646,0.12829594,0.0177804611,0.01734738226,0.008205907453,0.4598445173,0.06124391735,0.2497611901,0.0450396992,0.006108056089,0.0088521347,0.0030505955,0.003831411619
told_br_fine_grained,P0,4-shot,xlmr,all2,MDT-abl,0.04616593459,0.02595632218,0.08687937766,0.01841503414,0.08196676556,0.02229195931,0.01409994995,0.005024069595,0.2495142307,0.1160142256,0.1382175319,0.04492364461,0.0111093491,0.01575892685,0.02778098507,0.02006712317
told_br_fine_grained,P0,4-shot,xlmr,all2,-,0.02681318681,0.04831966974,0.04026838763,0.05052278702,0.08438589922,0.04002037943,0.02198814253,0.01859619065,0.4144124119,0.2626810627,0.0840866239,0.06920870633,0.000434310532,0.0008686210641,0.002698231246,0.002311472614
told_br_fine_grained,P0,4-shot,xlmr,all2_ab1,-,0.01592632022,0.01657267407,0.06222273581,0.0560254549,0.1058246929,0.01451141716,0.01973499048,0.008443481119,0.4716209782,0.06920453262,0.1583077413,0.04136594019,0.003586272775,0.003783840308,0.009373811639,0.01520409926
told_br_fine_grained,P0,8-shot,bert,base,-,0.02409481618,0.0148696826,0.05493382721,0.03087838882,0.09485103872,0.02833857912,0,0,0.4728238038,0.2168956144,0.103581517,0.02282343443,0.006699566603,0.005171240179,0.001823740275,0.002294687285
told_br_fine_grained,P0,8-shot,bert,sex_fine_grained1,-,0,0,0,0,0.136954929,0.000797933087,0.05259353807,0.004706462892,0.9060909652,0.0008945618629,0,0,0,0,0,0
told_br_fine_grained,P0,8-shot,bert,all2,-,0.01636651547,0.01671520535,0.03623226505,0.03868763521,0.1124099128,0.02081008585,0.01126414069,0.004413462864,0.6159193528,0.2279941662,0.08657642256,0.05194047553,0.0006980802792,0.001396160558,0.01981261245,0.02285702324
told_br_fine_grained,P0,8-shot,bert,all2_balanced,-,0.01513021223,0.005232862401,0.0983437924,0.03756259132,0.1115129598,0.01292581813,0.0140039423,0.009015689466,0.4788915332,0.07464681359,0.1466418603,0.03244001227,0.007786577821,0.005974692877,0.01979280008,0.01198033635
told_br_fine_grained,P0,8-shot,bert,all2_ab1,-,0.0170995298,0.0104588574,0.1130468053,0.05822656132,0.1101541519,0.02175598635,0.0126286277,0.00440064433,0.4343355047,0.05284471831,0.1492692125,0.07489179922,0.003358997993,0.001422954386,0.04134038518,0.02465693713
told_br_fine_grained,P0,8-shot,bert,all2_MLM,-,0.01328051531,0.01142142155,0.08657182215,0.02338604698,0.06936896886,0.02901842482,0.00435584772,0.002325569418,0.2309797685,0.2035064086,0.1340485202,0.01894208306,0.00923420468,0.01471793209,0.007112103404,0.01040143106
told_br_fine_grained,P0,8-shot,bert,all2_JOINT,-,0,0,0.007407407407,0,0.1000282748,0,0.002132196162,0,0.6880507578,0,0,0,0.00260756193,0,0,0
told_br_fine_grained,P0,8-shot,xlmr,base,LM-base,0.02654603199,0.00466158004,0.1379808619,0.03917144532,0.100716203,0.01946215698,0.01622042738,0.009891314775,0.3489473606,0.12975067,0.1638881798,0.01984242592,0.006867318856,0.008450607677,0.00456324059,0.003869839068
told_br_fine_grained,P0,8-shot,xlmr,all2,MDT,0.02968893367,0.01948915755,0.09201583112,0.07543342473,0.119224558,0.02514156828,0.02352710113,0.008288793667,0.4388080356,0.0839496237,0.2344860352,0.04815995698,0.006679370112,0.008224542778,0.009366599214,0.01102776957
\end{filecontents*}
\csvreader[range = {35,39}, late after line = \\]{results_told-br_fine_grained.csv}{}{
                    \myrow
                }
            \end{tabular}        }
        \caption{ToLD-Br fine-grained toxic Pt-Br}
            \end{subtable}
    \vspace{0.5cm}

    \hspace{10mm}
    \renewcommand\myrow{
    \csvcolvi & \perc{\csvcolvii} &
    \perc{\csvcolxi}
    }
    \begin{subtable}{0.28\linewidth}
        \resizebox{\textwidth}{!}{            \begin{tabular}{r|rr}
                               & hate & normal \\
                \hline
                \hline
                \begin{filecontents*}{results_stormfront_binary.csv}
,,,,,,hate_F1,hate_F1_STD,macro_averaged_F1,macro_averaged_F1_STD,noHate_F1,noHate_F1_STD
hate_speech18_binary,class,4-shot,bert,base,-,0.2471928014,0.073564958,0.5005771558,0.04553859429,0.7539615101,0.1138598986
hate_speech18_binary,class,4-shot,bert,fine2bin1,-,0.3261780657,0.01365314178,0.5813923042,0.02552034038,0.8366065427,0.05002971959
hate_speech18_binary,class,4-shot,bert,fine2bin1-MLM,-,0.1885059003,0.09692148619,0.5079709011,0.05242000401,0.8274359019,0.04892771486
hate_speech18_binary,class,4-shot,bert,fine2bin1-JOINT,-,0.1927175844,0,0.148575541,0,0.1044334975,0
hate_speech18_binary,class,4-shot,bert,fine2bin1_ab1,-,0.2474540714,0.09504550371,0.5134487079,0.1264006612,0.7794433443,0.1826997903
hate_speech18_binary,class,4-shot,bert,all1,-,0.3583927388,0.0385082156,0.6123162302,0.0323256155,0.8662397217,0.04565498512
hate_speech18_binary,class,4-shot,bert,all1_balanced,-,0.2835367623,0.0499047429,0.5084067256,0.08513007115,0.733276689,0.126277942
hate_speech18_binary,class,4-shot,bert,all1_MLM,-,0.1934717206,0.09915412961,0.495639672,0.06234240448,0.7978076235,0.06769891142
hate_speech18_binary,class,4-shot,bert,all1_JOINT,-,0.1903614458,0,0.4978458446,0,0.8053302433,0
hate_speech18_binary,class,4-shot,bert,all2,-,0.315527132,0.1030502202,0.5804956867,0.08107417988,0.8454642415,0.06866758643
hate_speech18_binary,class,4-shot,bert,all2_balanced,-,0.2329867663,0.06611191231,0.4693922578,0.08539238931,0.7057977493,0.1743155548
hate_speech18_binary,class,8-shot,bert,base,-,0.1863590428,0.1064452687,0.4918391007,0.04394511605,0.7973191586,0.1069199025
hate_speech18_binary,class,8-shot,bert,fine2bin1,-,0.3464692673,0.007576322906,0.5988340557,0.01520545532,0.8511988441,0.0284467209
hate_speech18_binary,class,8-shot,bert,fine2bin1_ab1,-,0.2412650668,0.1104003259,0.5263804613,0.09758974725,0.8114958559,0.1165528707
hate_speech18_binary,class,8-shot,bert,all1,-,0.3536640393,0.04533198822,0.6197258175,0.01581422921,0.8857875958,0.02369739965
hate_speech18_binary,class,8-shot,bert,all2,-,0.3594759591,0.03426646193,0.5997653812,0.05761670438,0.8400548033,0.08880067428
hate_speech18_binary,P0,zero_shot,bert,untrained,-,0.2087356322,0,0.1959672466,0,0.1831988609,0
hate_speech18_binary,P0,zero_shot,bert,fine2bin1,-,0.2325581395,0,0.558664243,0,0.8847703465,0
hate_speech18_binary,P0,zero_shot,bert,fine2bin1_ab1,-,0.2348008386,0,0.569437213,0,0.9040735874,0
hate_speech18_binary,P0,zero_shot,bert,all1,-,0.09210526316,0,0.5113618322,0,0.9306184012,0
hate_speech18_binary,P0,1-shot,bert,base,-,0.176256753,0.04474296905,0.4423872008,0.04902661343,0.7085176486,0.1336405809
hate_speech18_binary,P0,1-shot,bert,all2,-,0.3567562064,0.03341381958,0.583848453,0.06434083725,0.8109406997,0.09555105557
hate_speech18_binary,P0,1-shot,xlmr,base,LM-base,0.2631311814,0.006217313789,0.4653999575,0.0350951912,0.6676687336,0.06473688304
hate_speech18_binary,P0,1-shot,xlmr,all2,MDT,0.3502788316,0.05292211358,0.5880532137,0.07552449106,0.8258275958,0.1234833338
hate_speech18_binary,P0,4-shot,bert,base,-,0.2275350987,0.08017738853,0.4979629643,0.06732314573,0.76839083,0.1177111473
hate_speech18_binary,P0,4-shot,bert,fine2bin1,-,0.2949572422,0.04610424518,0.5648391382,0.02697617107,0.8347210342,0.06237662927
hate_speech18_binary,P0,4-shot,bert,fine2bin1_MLM,-,0.2259348386,0.06607001483,0.4891646024,0.01676667419,0.7523943662,0.06038873454
hate_speech18_binary,P0,4-shot,bert,fine2bin1_JOINT,-,0.3059760956,0,0.5091162275,0,0.7122563594,0
hate_speech18_binary,P0,4-shot,bert,fine2bin1_ab1,-,0.2619680331,0.03515773294,0.525777078,0.07083749091,0.789586123,0.1418837041
hate_speech18_binary,P0,4-shot,bert,all1,-,0.3304250087,0.0319788445,0.5912023146,0.01924231882,0.8519796206,0.05808261399
hate_speech18_binary,P0,4-shot,bert,all1_balanced,-,0.3405602225,0.01931810867,0.5800015441,0.03238797615,0.8194428657,0.05019845766
hate_speech18_binary,P0,4-shot,bert,all1_MLM,-,0.1879819798,0.09256511721,0.4824445838,0.006852828532,0.7769071878,0.0960356479
hate_speech18_binary,P0,4-shot,bert,all1_JOINT,-,0,0,0.4704427405,0,0.9408854811,0
hate_speech18_binary,P0,4-shot,bert,all2,-,0.3562263311,0.01889293196,0.610317902,0.02936804542,0.8644094729,0.05090163424
hate_speech18_binary,P0,4-shot,bert,all2_balanced,-,0.3471252422,0.0211329422,0.5899621127,0.04334803004,0.8327989833,0.06623908676
hate_speech18_binary,P0,4-shot,bert,all2_mlm_1_0,-,0.2947553867,0.06693333192,0.5655030715,0.01896810305,0.8362507563,0.06159829695
hate_speech18_binary,P0,4-shot,bert,all2_balanced_mlm_1_0,-,0.329059788,0.0165813896,0.566328563,0.03970651755,0.8035973381,0.07597954404
hate_speech18_binary,P0,4-shot,bert,all2_glob_balanced_mlm_1_0,-,0.2928366863,0.03771873507,0.5501066362,0.0340384821,0.807376586,0.07917665974
hate_speech18_binary,P0,4-shot,bert,all2_ab1,-,0.3458190583,0.01249426006,0.5921004345,0.01065919707,0.8383818107,0.02910041513
hate_speech18_binary,P0,4-shot,bert,all2_ab1_balanced,-,0.2937377475,0.01179136624,0.5641059804,0.02416398656,0.8344742134,0.05228869315
hate_speech18_binary,P0,4-shot,bert,all2_MLM,-,0.154632209,0.08026694555,0.4958157302,0.01181144833,0.8369992514,0.07031377588
hate_speech18_binary,P0,4-shot,bert,all2_JOINT,-,0,0,0.4692612791,0,0.9385225583,0
hate_speech18_binary,P0,4-shot,xlmr,base,LM-base,0.3027720129,0.04582158683,0.5281960896,0.06359568285,0.7536201663,0.09382839187
hate_speech18_binary,P0,4-shot,xlmr,all2,MLM,0.2545078988,0.1215492543,0.5488112037,0.04892900044,0.8431145086,0.09193305482
hate_speech18_binary,P0,4-shot,xlmr,all2,MTL,0.1630434783,0,0.5421758015,0,0.9213081247,0
hate_speech18_binary,P0,4-shot,xlmr,all2,Fusion,0.3018237214,0.0350925931,0.4679248636,0.07495145106,0.6340260058,0.115478792
hate_speech18_binary,P0,4-shot,xlmr,all2,MDT,0.368582741,0.02865686417,0.6040623856,0.05754208578,0.8395420302,0.09033858198
hate_speech18_binary,P0,4-shot,xlmr,all2,MDT-abl,0.2361214874,0.08764687421,0.5343499871,0.04439343655,0.8325784869,0.03925662442
hate_speech18_binary,P0,4-shot,xlmr,all2_ab1,-,0.3455862484,0.03171207452,0.6088432316,0.03632657298,0.8721002147,0.06931848568
hate_speech18_binary,P0,8-shot,bert,base,-,0.2323723549,0.05337150842,0.4856468248,0.04809047592,0.7389212946,0.08228509562
hate_speech18_binary,P0,8-shot,bert,fine2bin1,-,0.30934282,0.05797892368,0.5889821346,0.03033879294,0.8686214493,0.007319224497
hate_speech18_binary,P0,8-shot,bert,fine2bin1_ab1,-,0.2739056401,0.04520012301,0.5600467044,0.02467615454,0.8461877686,0.04777857574
hate_speech18_binary,P0,8-shot,bert,all1,-,0.3340263974,0.03370285782,0.5777585182,0.02374125322,0.8214906389,0.06373546434
hate_speech18_binary,P0,8-shot,bert,all2,-,0.3669604635,0.01445804958,0.6264284,0.009714692857,0.8858963365,0.01743492333
hate_speech18_binary,P0,8-shot,bert,all2_balanced,-,0.3573415172,0.007450326575,0.6023328197,0.01243717914,0.8473241223,0.02575175405
hate_speech18_binary,P0,8-shot,bert,all2_ab1,-,0.3539674424,0.02251178803,0.6032121173,0.01820063196,0.8524567921,0.02570879846
hate_speech18_binary,P0,8-shot,bert,all2_MLM,-,0.2045578914,0.06019389432,0.4701075998,0.04748822975,0.7356573082,0.103327481
hate_speech18_binary,P0,8-shot,bert,all2_JOINT,-,0.03100775194,0,0.4844402577,0,0.9378727634,0
hate_speech18_binary,P0,8-shot,xlmr,base,LM-base,0.3114492431,0.03755088366,0.5596379554,0.0309722481,0.8078266677,0.04962921106
hate_speech18_binary,P0,8-shot,xlmr,all2,MDT,0.3934430151,0.01675116505,0.6293063534,0.02701452706,0.8651696918,0.04132755347
hate_speech18_binary,P0,16-shot,bert,base,-,0.2899333538,0.03418600826,0.546520867,0.03618006391,0.8031083802,0.04414790216
hate_speech18_binary,P0,16-shot,bert,all2,-,0.3486068442,0.02976474651,0.600012276,0.06490511478,0.8514177078,0.1031564961
hate_speech18_binary,P0,16-shot,bert,all2_balanced,-,0.347215556,0.0280580674,0.5834229533,0.06331047949,0.8196303507,0.09938571014
hate_speech18_binary,P0,16-shot,xlmr,base,LM-base,0.348006363,0.04628846133,0.5931193344,0.04602079166,0.8382323058,0.07118603769
hate_speech18_binary,P0,16-shot,xlmr,all2,MDT,0.4022929484,0.02500892668,0.6398598447,0.0130820475,0.877426741,0.02169435991
hate_speech18_binary,P0,32-shot,bert,base,-,0.2750645437,0.03569799841,0.4948340143,0.07623258709,0.7146034849,0.13459069
hate_speech18_binary,P0,32-shot,bert,all2,-,0.367054188,0.02129877677,0.6281275189,0.02094047796,0.8892008498,0.0416216517
hate_speech18_binary,P0,32-shot,bert,all2_balanced,-,0.3644383438,0.01561816987,0.6040692554,0.02666223912,0.843700167,0.04149439069
hate_speech18_binary,P0,32-shot,xlmr,base,LM-base,0.3644917868,0.0495995287,0.5937207286,0.04986818002,0.8229496703,0.05439776539
hate_speech18_binary,P0,32-shot,xlmr,all2,MDT,0.4079021262,0.01417792921,0.6354921354,0.01287012766,0.8630821446,0.03151939865
hate_speech18_binary,P0,64-shot,bert,base,-,0.295417794,0.07133566199,0.5787962639,0.0530943945,0.8621747338,0.08888484204
hate_speech18_binary,P0,64-shot,bert,all2,-,0.3778836123,0.02519289346,0.6290273188,0.006638503432,0.8801710253,0.03208248921
hate_speech18_binary,P0,64-shot,bert,all2_balanced,-,0.371019625,0.01591065338,0.6169247401,0.007200757528,0.8628298553,0.01564461535
hate_speech18_binary,P0,64-shot,xlmr,base,LM-base,0.3878604804,0.07058283762,0.6174724024,0.05538846159,0.8470843245,0.0489700502
hate_speech18_binary,P0,64-shot,xlmr,all2,MDT,0.4038372619,0.01972812005,0.6365909884,0.02047301694,0.869344715,0.03584398528
hate_speech18_binary,P4,4-shot,bert,base,-,0.2150758836,0.06245728806,0.4941839179,0.06058623391,0.7732919522,0.07936780243
hate_speech18_binary,P5,4-shot,bert,base,-,0.1747521857,0.0708179324,0.5042919222,0.02369997388,0.8338316588,0.08265769474
\end{filecontents*}
\csvreader[range = {43,47}, late after line = \\]{results_stormfront_binary.csv}{}{
                    \myrow
                }
            \end{tabular}        }
        \caption{Stormfront binary hate En}
            \end{subtable}
    \hspace{15mm}
    \vspace{0.2cm}
                \renewcommand\myrow{
    \csvcolvi & \perc{\csvcolix} &
    \perc{\csvcolxi}
    }
    \begin{subtable}{0.30\linewidth}
        \resizebox{\textwidth}{!}{            \begin{tabular}{r|rr}
                          & misogyny & normal \\
                \hline
                \hline
                \begin{filecontents*}{results_ami_it-misogyny.csv}
,,,,,,macro_averaged_F1,macro_averaged_F1_STD,misogyny_F1,misogyny_F1_STD,non-misogyny_F1,non-misogyny_F1_STD
ami18_it-misogyny,class,4-shot,bert,base,-,0.6297925909,0.0414678705,0.6573348216,0.1033316707,0.6022503602,0.06617947502
ami18_it-misogyny,class,4-shot,bert,sexracistbin1,-,0.5514394842,0.06197457625,0.5318637382,0.1333110715,0.5710152302,0.05590135574
ami18_it-misogyny,class,4-shot,bert,sexracistbin1_MLM,-,0.7117487342,0.006134909444,0.7356155784,0.03079964024,0.6878818901,0.03211039686
ami18_it-misogyny,class,4-shot,bert,sexracistbin1_JOINT,-,0.4197802198,0,0.5938461538,0,0.2457142857,0
ami18_it-misogyny,class,4-shot,bert,sexracistbin1_ab1,-,0.6048413376,0.03742330832,0.5974362879,0.057136463,0.6122463874,0.06919602316
ami18_it-misogyny,class,4-shot,bert,sexracistbin2,-,0.6574874636,0.07643652425,0.6512155297,0.05964268808,0.6637593975,0.09845476809
ami18_it-misogyny,class,4-shot,bert,sexracistbin2_MLM,-,0.728745171,0.03066850792,0.7407553583,0.01048340924,0.7167349836,0.05592197744
ami18_it-misogyny,class,4-shot,bert,sexracistbin2_JOINT,-,0.4516681092,0,0.688547486,0,0.2147887324,0
ami18_it-misogyny,class,4-shot,bert,sexracistbin2_ab1,-,0.6583984123,0.09727589372,0.6443329699,0.1343638792,0.6724638547,0.06455548189
ami18_it-misogyny,class,4-shot,bert,all1,-,0.638909947,0.05475737781,0.6254678834,0.06682727527,0.6523520106,0.06362473952
ami18_it-misogyny,class,4-shot,bert,all1_MLM,-,0.703435791,0.03130590932,0.7482079215,0.01455080016,0.6586636605,0.04924420118
ami18_it-misogyny,class,4-shot,bert,all1_JOINT,-,0.4298081827,0,0.4651600753,0,0.39445629,0
ami18_it-misogyny,class,4-shot,bert,all2,-,0.6318651097,0.04377529155,0.6156424884,0.07260907562,0.6480877309,0.04974944064
ami18_it-misogyny,class,4-shot,bert,all2_balanced,-,0.6838165261,0.06293513605,0.6625601536,0.07994320277,0.7050728986,0.05021969651
ami18_it-misogyny,class,8-shot,bert,base,-,0.6325727154,0.1078378237,0.6068933998,0.1459652996,0.658252031,0.07399797634
ami18_it-misogyny,class,8-shot,bert,sexracistbin1,-,0.6004901565,0.03029102903,0.5989059151,0.04764959897,0.6020743979,0.04747784179
ami18_it-misogyny,class,8-shot,bert,sexracistbin1_ab1,-,0.64791141,0.01932734974,0.6597562356,0.04277431217,0.6360665844,0.03257729071
ami18_it-misogyny,class,8-shot,bert,sexracistbin2,-,0.6801763349,0.06659533054,0.681237139,0.1026489182,0.6791155307,0.03540071999
ami18_it-misogyny,class,8-shot,bert,sexracistbin2_ab1,-,0.5983128197,0.108044972,0.576113239,0.1570933739,0.6205124003,0.06024410086
ami18_it-misogyny,class,8-shot,bert,all1,-,0.6654554769,0.04991649113,0.6515478291,0.09497397787,0.6793631247,0.02899202674
ami18_it-misogyny,class,8-shot,bert,all2,-,0.664109229,0.02735637406,0.6470098839,0.04996652386,0.6812085741,0.01804162673
ami18_it-misogyny,P4,zero_shot,bert,untrained,-,0.3426962306,0,0.6733200266,0,0.01207243461,0
ami18_it-misogyny,P4,zero_shot,bert,sexracistbin1,-,0.4352124924,0,0.2138157895,0,0.6566091954,0
ami18_it-misogyny,P4,zero_shot,bert,sexracistbin1_ab1,-,0.4342663627,0,0.2458770615,0,0.6226556639,0
ami18_it-misogyny,P4,zero_shot,bert,sexracistbin2,-,0.3481985692,0,0.03468208092,0,0.6617150574,0
ami18_it-misogyny,P4,zero_shot,bert,sexracistbin2_ab1,-,0.3449045496,0,0.03766478343,0,0.6521443159,0
ami18_it-misogyny,P4,zero_shot,bert,all1,-,0.4201575807,0,0.1957585644,0,0.644556597,0
ami18_it-misogyny,P4,1-shot,bert,base,-,0.476347172,0.08234986016,0.5935786745,0.1837687252,0.3591156694,0.1501517056
ami18_it-misogyny,P4,1-shot,bert,all2,-,0.5808876862,0.08242884428,0.5618677266,0.1600129582,0.5999076459,0.07928924883
ami18_it-misogyny,P4,1-shot,xlmr,base,LM-base,0.5801656137,0.08470205237,0.563715237,0.176981291,0.5966159905,0.115311534
ami18_it-misogyny,P4,1-shot,xlmr,all2,MDT,0.4943755845,0.04028202496,0.5309994599,0.08665797888,0.4577517091,0.1343914369
ami18_it-misogyny,P4,4-shot,bert,base,-,0.6125772613,0.09793406241,0.6649730295,0.1176921118,0.5601814932,0.1016414839
ami18_it-misogyny,P4,4-shot,bert,sexracistbin1,-,0.6184742928,0.02689080153,0.6113770955,0.043222384,0.6255714901,0.05685724037
ami18_it-misogyny,P4,4-shot,bert,sexracistbin1_MLM,-,0.6279911345,0.1080392571,0.6347869276,0.1419675435,0.6211953415,0.0830485717
ami18_it-misogyny,P4,4-shot,bert,sexracistbin1_JOINT,-,0.4264294743,0,0.2595204513,0,0.5933384973,0
ami18_it-misogyny,P4,4-shot,bert,sexracistbin1_ab1,-,0.5576386414,0.06829143767,0.4855352483,0.1325778714,0.6297420345,0.03042456942
ami18_it-misogyny,P4,4-shot,bert,sexracistbin2,-,0.6873535622,0.04348937799,0.7344525708,0.01890669891,0.6402545536,0.06932524321
ami18_it-misogyny,P4,4-shot,bert,sexracistbin2_MLM,-,0.617429469,0.06927677018,0.6690071512,0.0898352102,0.5658517869,0.08780884605
ami18_it-misogyny,P4,4-shot,bert,sexracistbin2_JOINT,-,0.3293091885,0,0,0,0.6586183769,0
ami18_it-misogyny,P4,4-shot,bert,sexracistbin2_ab1,-,0.4805288422,0.08364793014,0.3966154381,0.2235291534,0.5644422463,0.08645887013
ami18_it-misogyny,P4,4-shot,bert,all1,-,0.6591306556,0.04287209924,0.6380161434,0.08422432155,0.6802451678,0.02200151551
ami18_it-misogyny,P4,4-shot,bert,all1_MLM,-,0.615829158,0.09676452259,0.6327123013,0.1811358476,0.5989460146,0.04632830453
ami18_it-misogyny,P4,4-shot,bert,all1_JOINT,-,0.4569733917,0,0.4531722054,0,0.460774578,0
ami18_it-misogyny,P4,4-shot,bert,all2,-,0.6591306556,0.04287209924,0.6380161434,0.08422432155,0.6802451678,0.02200151551
ami18_it-misogyny,P4,4-shot,bert,all2_balanced,-,0.70305033,0.01391561847,0.7132319794,0.02642537236,0.6928686806,0.02315745592
ami18_it-misogyny,P4,4-shot,bert,all2_glob_balanced_mlm_1_0,-,0.6140816762,0.03398688483,0.6404851064,0.05217506694,0.587678246,0.09567364948
ami18_it-misogyny,P4,4-shot,bert,all2_ab1,-,0.6416453361,0.06088237648,0.6362858205,0.107226827,0.6470048516,0.05892595301
ami18_it-misogyny,P4,4-shot,bert,all2_ab1_balanced,-,0.6317189346,0.06063349566,0.6148997372,0.07550174342,0.6485381321,0.05573306754
ami18_it-misogyny,P4,4-shot,bert,all2_MLM,-,0.615829158,0.09676452259,0.6327123013,0.1811358476,0.5989460146,0.04632830453
ami18_it-misogyny,P4,4-shot,bert,all2_JOINT,-,0.4569733917,0,0.4531722054,0,0.460774578,0
ami18_it-misogyny,P4,4-shot,bert,all2_cross,-,0.579410997,0.05522863193,0.534820964,0.09603288134,0.6240010299,0.05765414378
ami18_it-misogyny,P4,4-shot,bert,all2_cross2,-,0.585124631,0.05673332017,0.5816395189,0.07797148785,0.588609743,0.08444758468
ami18_it-misogyny,P4,4-shot,bert,all2_cross3,-,0.6249861951,0.06663708439,0.5938333976,0.1292712332,0.6561389926,0.02853315596
ami18_it-misogyny,P4,4-shot,bert,all2_cross4,-,0.6576590756,0.05547684428,0.6913073688,0.02647268276,0.6240107823,0.08912552903
ami18_it-misogyny,P4,4-shot,xlmr,base,LM-base,0.6033058921,0.1044658662,0.5526331473,0.2134330678,0.6539786369,0.04064788408
ami18_it-misogyny,P4,4-shot,xlmr,all2,MLM,0.642433138,0.05818157992,0.6923286778,0.05453895183,0.5925375983,0.1088705791
ami18_it-misogyny,P4,4-shot,xlmr,all2,MTL,0.4532333244,0,0.4078516903,0,0.4986149584,0
ami18_it-misogyny,P4,4-shot,xlmr,all2,Fusion,0.3911368718,0.1116979191,0.5350633426,0.267681988,0.247210401,0.3038561665
ami18_it-misogyny,P4,4-shot,xlmr,all2,MDT,0.6698495591,0.04029279349,0.6932610507,0.05389453399,0.6464380675,0.05098426094
ami18_it-misogyny,P4,4-shot,xlmr,all2,MDT-abl,0.6544588204,0.04023430502,0.6653935337,0.07192835836,0.6435241072,0.05031813034
ami18_it-misogyny,P4,4-shot,xlmr,all2,MDT-joint,0.6671154114,0.05864391649,0.6656364356,0.08559035615,0.6685943872,0.06990090141
ami18_it-misogyny,P4,4-shot,xlmr,all2,MDT-3-steps,0.5866304895,0.1041549963,0.5260851903,0.1787334846,0.6471757887,0.04339495956
ami18_it-misogyny,P4,4-shot,xlmr,all2,MDT-single,0.6467812531,0.09943097001,0.6778174719,0.1211559102,0.6157450342,0.1541046126
ami18_it-misogyny,P4,4-shot,xlmr,all2_ab1,-,0.6804463189,0.05430078024,0.6816498196,0.0879696589,0.6792428182,0.06228726587
ami18_it-misogyny,P4,8-shot,bert,base,-,0.6266834856,0.051340283,0.6021620604,0.1096365788,0.6512049109,0.04310296602
ami18_it-misogyny,P4,8-shot,bert,sexracistbin1,-,0.6065046297,0.05447540805,0.5716289106,0.1143936583,0.6413803488,0.02242279225
ami18_it-misogyny,P4,8-shot,bert,sexracistbin1_ab1,-,0.5277802499,0.0779632403,0.422261666,0.1394567577,0.6332988339,0.03345546265
ami18_it-misogyny,P4,8-shot,bert,sexracistbin2,-,0.6327058299,0.05130209809,0.6624623317,0.07467084566,0.602949328,0.07450090194
ami18_it-misogyny,P4,8-shot,bert,sexracistbin2_ab1,-,0.5772617416,0.04128779866,0.6105217188,0.07448380863,0.5440017644,0.05403146241
ami18_it-misogyny,P4,8-shot,bert,all1,-,0.6297219609,0.01997982916,0.5874885725,0.06346336475,0.6719553493,0.05300277725
ami18_it-misogyny,P4,8-shot,bert,all2,-,0.6297219609,0.01997982916,0.5874885725,0.06346336475,0.6719553493,0.05300277725
ami18_it-misogyny,P4,8-shot,bert,all2_ab1,-,0.6819253888,0.02868423828,0.6625617705,0.04460316017,0.701289007,0.02031676667
ami18_it-misogyny,P4,8-shot,bert,all2_MLM,-,0.6741946999,0.04444453371,0.6806116973,0.06755309746,0.6677777025,0.06673400567
ami18_it-misogyny,P4,8-shot,bert,all2_JOINT,-,0.5546349774,0,0.5105438402,0,0.5987261146,0
ami18_it-misogyny,P4,8-shot,xlmr,base,LM-base,0.6549258991,0.03169518757,0.7077144436,0.03417300117,0.6021373546,0.05119134615
ami18_it-misogyny,P4,8-shot,xlmr,all2,MDT,0.6729926898,0.06462037914,0.6895292878,0.08800531444,0.6564560918,0.06411708946
ami18_it-misogyny,P4,16-shot,bert,base,-,0.704600034,0.03457327958,0.7380613864,0.02576756202,0.6711386816,0.0501125053
ami18_it-misogyny,P4,16-shot,bert,all2,-,0.6997972498,0.03092605577,0.7070274218,0.02611053187,0.6925670777,0.04045932108
ami18_it-misogyny,P4,16-shot,xlmr,base,LM-base,0.7205474482,0.05731645237,0.7411183096,0.02373366949,0.6999765867,0.1027515811
ami18_it-misogyny,P4,16-shot,xlmr,all2,MDT,0.7377960129,0.02978155779,0.7581608084,0.03341625342,0.7174312173,0.03272152515
ami18_it-misogyny,P4,32-shot,bert,base,-,0.7402359674,0.02626526288,0.7784251798,0.02005004328,0.702046755,0.04817290128
ami18_it-misogyny,P4,32-shot,bert,all2,-,0.7229038408,0.01769068578,0.7390731434,0.0138904606,0.7067345382,0.02941793432
ami18_it-misogyny,P4,32-shot,xlmr,base,LM-base,0.7558991243,0.06123076484,0.7782889191,0.06626549252,0.7335093294,0.05901896016
ami18_it-misogyny,P4,32-shot,xlmr,all2,MDT,0.760533315,0.02524845327,0.7833463897,0.01542743544,0.7377202404,0.04599240116
ami18_it-misogyny,P4,64-shot,bert,base,-,0.7512732967,0.03817088081,0.7668591613,0.04467637367,0.7356874321,0.05929058712
ami18_it-misogyny,P4,64-shot,bert,all2,-,0.7453360339,0.02180146232,0.7619273324,0.01647233862,0.7287447354,0.02826116178
ami18_it-misogyny,P4,64-shot,xlmr,base,LM-base,0.7899825909,0.02841677458,0.8123169956,0.01859287032,0.7676481862,0.03940197419
ami18_it-misogyny,P4,64-shot,xlmr,all2,MDT,0.7760692622,0.02652618572,0.8040484417,0.01222874651,0.7480900827,0.04187457149
\end{filecontents*}
\csvreader[range = {55,59}, late after line = \\]{results_ami_it-misogyny.csv}{}{
                    \myrow
                }
            \end{tabular}        }
        \caption{AMI binary misogyny It}
            \end{subtable}
    \vspace{0.1cm}
        \caption{
        Per label 4-shot $F_1$ scores (\%).
        Unseen labels are \textbf{bolded}.
    }
    \label{tab:per_label_results}
\end{table*}

Although all labels of the binary target datasets are seen, as mentioned the
definitions
of some labels are different.
For example, the offensive label of the \texttt{OLID binary offensive En}
target dataset includes profanity,
while the same label in the external \texttt{HASOC fine-grained
abusive En} dataset does not.
However, due to the inclusion of external training samples that are directly
labeled as profane, the model is trained on all the necessary information.
It only has to learn to combine them in the final model of a given target
dataset, such as profanity and the more restrictive offensive label of \texttt{HASOC}
into the general offensive label of \texttt{OLID}.

All the used external datasets are English.
Comparing the improvements of the monolingual and cross-lingual setups of MDT,
i.e., English and non-English target datasets, we found that the external
datasets are more beneficial monolingually.
The average improvements are 9.51\% (ignoring \texttt{AMI misogyny En}) and
6.09\% respectively.
This is not surprising given that cross-lingual transfer learning is almost
always less effective.
Still, it shows that the combination with English external
datasets
is beneficial to non-English test corpora as well.
This is an important use
case for reducing costs by
dramatically reducing the need for human annotation.
Additionally, all the external datasets contain Twitter or Facebook posts,
while the Stormfront target dataset contains forum posts which tend to be longer
and have different language use compared to microblog posts.
Even in this case, the improvements of MDT
are large compared
with the baseline showing the generality of our model.

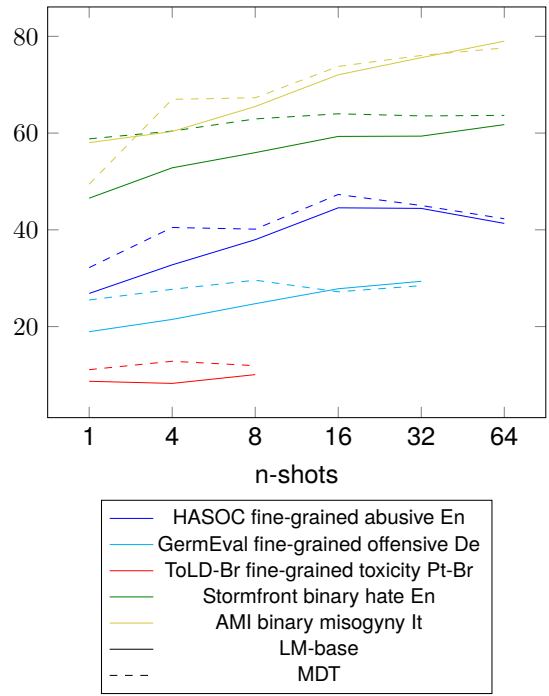
\begin{figure}[t]
    \begin{center}
    \resizebox{.95\columnwidth}{!}{        \begin{tikzpicture}
            \begin{axis}[
                xtick=data,
                xticklabels={1, 4, 8, 16, 32, 64},
                xmode=log,
                xlabel={n-shots},
                                legend style={
                    at={(0.5,-0.20)},
                    anchor=north,
                    nodes={scale=0.8, transform shape}
                }
            ]
                                \addplot[solid, color=blue, mark=none] table [
                    x=n,
                    y=HASOC fine-grained abusive En LM-base,
                    col sep=comma
                ] {n-shot.csv};
                \addplot[dashed, color=blue, mark=none] table [
                    x=n,
                    y=HASOC fine-grained abusive En MDT,
                    col sep=comma,
                    forget plot
                ] {n-shot.csv};
                \addlegendentry{HASOC fine-grained abusive En};
                
                                \addplot[solid, color=cyan, mark=none] table [
                    x=n,
                    y=GermEval fine-grained offensive De LM-base,
                    col sep=comma
                ] {n-shot.csv};
                \addplot[dashed, color=cyan, mark=none] table [
                    x=n,
                    y=GermEval fine-grained offensive De MDT,
                    col sep=comma,
                    forget plot
                ] {n-shot.csv};
                \addlegendentry{GermEval fine-grained offensive De};
                
                                \addplot[solid, color=red, mark=none] table [
                    x=n,
                    y=ToLD-Br fine-grained toxicity Pt-Br LM-base,
                    col sep=comma
                ] {n-shot.csv};
                \addplot[dashed, color=red, mark=none] table [
                    x=n,
                    y=ToLD-Br fine-grained toxicity Pt-Br MDT,
                    col sep=comma,
                    forget plot
                ] {n-shot.csv};
                \addlegendentry{ToLD-Br fine-grained toxicity Pt-Br};
                
                                \addplot[solid, color=green!50!black, mark=none] table [
                    x=n,
                    y=Stormfront binary hate En LM-base,
                    col sep=comma
                ] {n-shot.csv};
                \addplot[dashed, color=green!50!black, mark=none] table [
                    x=n,
                    y=Stormfront binary hate En MDT,
                    col sep=comma,
                    forget plot
                ] {n-shot.csv};
                \addlegendentry{Stormfront binary hate En};
                
                                \addplot[solid, color=yellow!80!black, mark=none] table [
                    x=n,
                    y=AMI binary misogyny It LM-base,
                    col sep=comma
                ] {n-shot.csv};
                \addplot[dashed, color=yellow!80!black, mark=none] table [
                    x=n,
                    y=AMI binary misogyny It MDT,
                    col sep=comma,
                    forget plot
                ] {n-shot.csv};
                \addlegendentry{AMI binary misogyny It};
                
                \addlegendimage{}
                \addlegendentry{LM-base};
                \addlegendimage{dashed}
                \addlegendentry{MDT};
            \end{axis}
        \end{tikzpicture}
    }
    \end{center}
    \caption{
        $F_1$ scores (\%) at different number (n-shots) of target dataset
        samples comparing LM-base with our MDT approach.
        Some datasets have less than $64$ samples available for the minority
        label.
    }
    \label{fig:nshot}
\end{figure}

\paragraph{N-shot analysis}
Due to the high label bias of abusive language datasets, we consider 4-shots of
the target datasets for efficiency.
For example, out of the $12\,833$ training samples in the ToLD-Br dataset only 11
(0.08\%) are labeled as racism, meaning that a given annotator has to look at
more than a thousand text inputs to increase the number of this minority label
with just one sample.
For a more complete picture of the performance of our approach however, we
present experiments
with
different
$n$ values on a few selected datasets in
Figure~\ref{fig:nshot},
while results on all the datasets are in Table~\ref{tab:all_results} in the appendix.
Similarly to \citet{zhao-etal-2021-closer} we find that although
we average our results over 5 seeds, the performance can be unstable at lower
$n$ values.
It even decreases with the increase of training samples in same
cases.
However, MDT
steadily
outperforms LM-base, the gap only decreases at higher
$n$ values.
Still, even at $n=64$ the baseline performs worse on
2 of the 3 datasets.
In contrast, MDT has the largest improvements compared to the
baseline at
lower $n$ values,
which shows the strong advantage of using external datasets,
especially for target datasets
such as \texttt{ToLD-Br}, for which
acquiring even one
racism sample
is expensive.

\paragraph{Per label analysis}
We present per label $F_1$ scores on a few selected datasets in
Table~\ref{tab:per_label_results}.
In case of the fine-grained \texttt{HASOC} dataset in subtable (a), the label
\emph{hate} was significantly improved, while the performance on
\emph{offensive} and \emph{profane} were improved and decreased respectively
with a similar margin.
All labels were improved on the \texttt{OLID} dataset in subtable (b), even the
unseen \emph{other} label by almost 5 percentage points.
Most interestingly, 3 out 4 unseen labels (5 out of 7 overall) were improved on the
\texttt{ToLD-Br} dataset in subtable (c).
Our conjecture is that the unseen \emph{insult} label is related to the
\emph{fearful} label of the external \texttt{MLMA} dataset,
since texts
causing fear often involve insults as well, which leads to
this improvement.
Additionally,
as stated by the authors of \texttt{ToLD-Br} \cite{leite-etal-2020-toxic}, the
unseen \emph{insult} and \emph{obscene} labels were often confused by
the annotators, indicating their similarity, thus the latter could have also
benefited from the \emph{fearful} instances of \texttt{MLMA}.
Similarly, the \emph{LMBTQ+phobia} label is to some extent related to
\emph{sexism} external instances, thus MDT can leverage their similarity
automatically without the need for manual label modifications.
In contrast, \emph{xenophobia}, which is somewhat related to \emph{racism}, was
not improved.
However,
since both models achieve
less than $1$ percent $F1$, we believe that \emph{xenophobia}
is simply too hard to classify, which is the reason for no improvements in MDT.
On the binary target datasets in subtables (d) and (e) the results are similar
as in case of the fine-grained datasets, however all labels were improved
in
\texttt{Stormfront}, while only \emph{misogyny} improved
in
\texttt{AMI}.

\paragraph{Ablation study}
We were interested in the
additional value of labels which are not directly used in the target datasets.
Thus,
we removed the external-only
labels from the external datasets in step 1,
i.e., all labels which are not part of
a given target dataset,
and performed step 2 as normal (MDT\textsubscript{abl}).
We present experiments on datasets where MDT outperformed LM-base in
Table~\ref{tab:ablation_result_multi}.
In the majority of the cases, especially for the binary datasets, we found that
removing labels from the external datasets deteriorates the model's performance
supporting our claim that by training first on a large set of diverse datasets
the model can learn a general knowledge of abusive language which is then
beneficial for the target task.
Even though some labels in the external datasets aren't directly used for
a given target task they are still beneficial.

\begin{table}[t]
    \newcommand\myrow{
    \csvcoli & \csvcolii & \csvcoliii & \csvcoliv
    }
    \centering
    \resizebox{.90\columnwidth}{!}{        \begin{tabular}{rr|rr}
        & & MDT & MDT\textsubscript{abl}           \\
        \hline
        \hline
        \begin{filecontents*}{results_ablation.csv}
,,MDT,ablation
\multirow{6}{*}{\rotatebox{90}{fine-grained}},HASOC abusive En,\textbf{40.48},36.96
,HASOC abusive Hi,34.36,\textbf{38.38}
,HASOC abusive De,\textbf{33.96},32.55
,GermEval offensive De,27.70,\textbf{28.04}
,ToLD-Br toxicity Pt-Br,\textbf{12.83},8.20
,OLID target EN,\textbf{49.55},44.73
\hline\multirow{6}{*}{\rotatebox{90}{binary}},Stormfront hate En,\textbf{60.41},53.43
,HatEval hate En,\textbf{60.20},56.91
,HatEval hate Es,\textbf{54.47},52.69
,OLID offensive En,\textbf{64.81},45.09
,GermEval offensive De,\textbf{65.02},56.95
,AMI misogyny It,\textbf{66.98},65.45
\hline,avg.,\textbf{47.56},43.28
\end{filecontents*}
\csvreader[
            late after line = \\
        ]{results_ablation.csv}{}{
            \myrow
        }
        \end{tabular}    }
    \caption{
        Macro averaged 4-shot $F_1$ scores of our
        ablation
                study which have labels
        removed from the external datasets that are not needed in the target
        datasets (MDT\textsubscript{abl}).
    }
    \label{tab:ablation_result_multi}
        \end{table}

\begin{table}[t]
    \newcommand\myrow{
    \csvcoli & \csvcolii & \csvcoliii
    }
    \centering
            \begin{tabular}{r|rr}
        & LM-base & MDT \\
        \hline
        \hline
        \begin{filecontents*}{results_hatecheck.csv}
,LM-base,MDT
F1-4,45.23,\textbf{61.95}
F5-6,48.59,\textbf{49.72}
F7-9,28.88,\textbf{30.36}
F10-11,50.00,\textbf{52.38}
F12-13,\textbf{74.91},74.73
F14-15,\textbf{29.15},24.89
F16-17,\textbf{73.18},49.53
F18-19,\textbf{27.14},9.57
F20-21,\textbf{8.30},4.44
F22-24,14.51,\textbf{21.82}
F25-29,41.46,\textbf{59.44}
\end{filecontents*}
\csvreader[
                        late after line = \\
        ]{results_hatecheck.csv}{}{
            \myrow
        }
        \end{tabular}        \caption{
    Macro averaged 4-shot $F_1$ scores on the \textsc{HateCheck} functional
    test classes using \texttt{HatEval binary hate En} as the target dataset.
    The results are averaged over test cases in the classes.
    The naming F$X-Y$ refers to the test class containing test cases between $X$ and $Y$.
    }
    \label{tab:hatecheck}
\end{table}

\paragraph{\textsc{HateCheck}}
is a test suite containing 29 functional tests grouped into
11 classes \cite{rottger-etal-2021-hatecheck}.
It focuses on testing various aspects of hate speech detection models.
It defines the following test classes: derogation (\textbf{F1-4}), threatening
language (\textbf{F5-6}), slur usage (\textbf{F7-9}), profanity usage
(\textbf{F10-11}), profane reference (\textbf{F12-13}), negation (\textbf{F14-15}),
paraphrasing (\textbf{F16-17}), non-hate when mentioning protected groups
(\textbf{F18-19}), counter speech (\textbf{F20-21}), abuse against non-protected
targets (\textbf{F22-24}) and spelling variations (\textbf{F25-29}).
We evaluated MDT using \texttt{HatEval binary hate En} as the target dataset
compared to LM-base in Table~\ref{tab:hatecheck}.
We found that MDT improves on cases testing a higher level of hate speech
understanding, such as derogation and threatening, or slur and profanity usage,
while the performance decreases on cases testing
linguistics phenomena, such as negation or paraphrasing.
These results further support our claim that MDT results in models with better
abusive language understanding.
On the other hand, they also highlight sensitivity to linguistic phenomena which
should be improved in future work.

\section{Conclusions}
\label{sec:conclusions}

Due to the large variety of the abusive content to be filtered, lack of resources
is a major problem, larger than for many other NLP tasks.
In order to eliminate the need for expensive dataset annotation for novel
application scenarios,
and
thus reduce costs, we proposed a two-step multi-dataset
approach (MDT) which exploits datasets we already have to learn general abusive
language understanding and requires only a few annotated samples for the target
task.
Our experiments on various datasets showed that external datasets improve
few-shot classification across tasks, text genres and languages.
Additionally, our analysis reveals that
external-only labels also contain useful information and
consequently, that
unseen labels
can be improved
as well,
arguing for the utility of our approach.

\section*{Limitations}
\label{sec:limitations}

The aim of our approach is that given the requirements (labels and their
definitions) of a target task, train an abusive language classifier using
external datasets.
Although
our approach supports selecting more closely related datasets, e.g.,
only hate speech or only misogyny and sexism datasets, we used a broad range of
abusive language related external datasets in our experiments for simplicity, and
to save computational resources.
On the one hand, this setup shows the general applicability of MDT.
However, our
results do not show the potential of more specialized (but still easy to set up)
configurations.

Additionally, we only used English corpora as external datasets.
Although
English has the most abusive language resources, datasets of
other languages could also be used for this purpose.
Our results show that MDT is more beneficial monolingually than
cross-lingually, thus using resources from the same language as a non-English
target dataset could be beneficial, which
we did not test in this work.

Finally, our
analysis
on HateCheck
reveals some weaknesses of our approach which
indicate future directions in combining abusive language datasets.

\section*{Acknowledgements}
\label{sec:ack}

We thank the anonymous reviewers for their helpful feedback.
The work was
funded by the European Research Council
(ERC; grant agreement No.~640550) and by the German Research Foundation (DFG;
grant FR 2829/4-1 and FR 2829/7-1).

\section{Bibliographical References}\label{reference}

\bibliographystyle{lrec_natbib}
\bibliography{custom,anthology}

\appendix

\section{Additional Details}
\label{sec:appendix_details}

We present details of the used datasets in Table~\ref{tab:datasets}, such as
source platform, number of samples (we only used 4 samples per label for
training in the main experiments and an overall 16 samples following the
original label distribution for
validation in case of the target datasets) and used PVPs.
We kept PVPs simple and uniform across datasets using English PVPs even for
non-English datasets as well.
We note however that in our initial experiments we tested machine translated
PVPs which did not lead to significantly different results
\cite{zhao-schutze-2021-discrete}.
Additionally, if a given token related to a label is split by the tokenizer,
e.g., \emph{dominance} $\rightarrow$ [\emph{domina}, \emph{\#nce}], we take the
averaged probabilities of the subwords at the \texttt{[MASK]} position as the
probability of the related label.
Finally, we show complete results of all setups in
Table~\ref{tab:all_results} (containing each target dataset in a separate
subtable stretching over multiple pages).

\begin{table*}[t]
    \centering
    \resizebox{.95\textwidth}{!}{        \begin{tabular}{r|lrrrll}
                                                & source                               &
            \#train                             & \#valid                              & \#test & verbalizer

                                                & Pattern
            \\
            \hline
            \hline
            AMI fine-grained misogyny En        & Twitter                              &
            1,428                               & 357                                  & 460    & \makecell[l]{ stereotypical $\rightarrow$ stereotype
            \\ dominance $\rightarrow$ dominance \\ derailing $\rightarrow$ derailing \\
            harassment $\rightarrow$ sexual\_harassment                                                                                                                                     \\ discrediting $\rightarrow$
            discredit}
            & \texttt{X}
            $\rightarrow$ \texttt{X} It was \texttt{{[}MASK{]}}                                                                                                                                               \\
            \hline
            AMI binary misogyny En              & Twitter                              &
            3,200                               & 800 & 1,000  & \makecell[l]{ sexist $\rightarrow$ misogyny                                                 \\ neutral
            $\rightarrow$ normal}

                                                & \texttt{X} $\rightarrow$
                                                \texttt{X}
                                                It was \texttt{{[}MASK{]}}                                                                                                                                                               \\
            \hline
            AMI binary misogyny It              & Twitter                              &
            3,200                               & 800                                  & 1,000  & \makecell[l]{ sexist $\rightarrow$ misogyny                                                 \\ neutral
            $\rightarrow$ normal}

                                                & \texttt{X} $\rightarrow$
                                                \texttt{X}
                                                It was \texttt{{[}MASK{]}}                                                                                                                                                               \\
                                                                                    \hline
            GermEval fine-grained offensive De  & Twitter                              &
            4,007                               & 1,002                                & 3,532  & \makecell[l]{ profane $\rightarrow$ profanity                                             \\
            insulting $\rightarrow$ insult                                                                                                                                                  \\ abusive $\rightarrow$ abusive \\ neutral
            $\rightarrow$ normal}

                                                & \texttt{X} $\rightarrow$
                                                \texttt{X} It was
                                                \texttt{{[}MASK{]}}                                                                                                       \\
            \hline
            GermEval binary offensive De        & Twitter                              &
            4,007                               & 1,002                                & 3,532  & \makecell[l]{ offensive $\rightarrow$ offensive                                           \\
            neutral $\rightarrow$ normal}

                                                & \texttt{X} $\rightarrow$
                                                \texttt{X} It was
                                                \texttt{{[}MASK{]}}                                                                                                                                                             \\
            \hline
            HASOC fine-grained abusive En         & Twitter, Facebook                    &
            1,808                               & 453                                  & 288    & \makecell[l]{ hate $\rightarrow$ hate                                                     \\ offensive
            $\rightarrow$ offensive                                                                                                                                                         \\ profane $\rightarrow$ profanity}

                                                & \texttt{X} $\rightarrow$
                                                \texttt{X} It was
                                                \texttt{{[}MASK{]}}
            \\
            \hline
            HASOC fine-grained abusive De         & Twitter, Facebook                    &
            325                                 & 82                                   & 136    & \makecell[l]{ hate $\rightarrow$ hate                                                     \\ offensive
            $\rightarrow$ offensive                                                                                                                                                         \\ profane $\rightarrow$ profanity}

                                                & \texttt{X} $\rightarrow$
                                                \texttt{X} It was
                                                \texttt{{[}MASK{]}}
            \\
            \hline
            HASOC fine-grained abusive Hi         & Twitter, Facebook                    &
            1,975                               & 494                                  & 605    & \makecell[l]{ hate $\rightarrow$ hate                                                     \\ offensive
            $\rightarrow$ offensive                                                                                                                                                         \\ profane $\rightarrow$ profanity}

                                                & \texttt{X} $\rightarrow$
                                                \texttt{X} It was
                                                \texttt{{[}MASK{]}}
            \\
            \hline
            HatEval binary hate En              & Twitter                              &
            3,055                               & 764                                  & 850    & \makecell[l]{ hate $\rightarrow$ hateful                                                     \\ neutral
            $\rightarrow$ normal}

                                                & \texttt{X}
                                                $\rightarrow$ \texttt{X} It was
                                                \texttt{{[}MASK{]}}\\
            \hline
            HatEval binary hate Es              & Twitter                              &
            3,560                              & 890
            & 500    & \makecell[l]{ hate $\rightarrow$ hateful                                                     \\ neutral
            $\rightarrow$ normal}

                                                & \texttt{X}
                                                $\rightarrow$ \texttt{X} It was
                                                \texttt{{[}MASK{]}}\\
            \hline
            LSA fine-grained abusive En         & Twitter                              &
            29,728                              & 7,433                                & 9,291  & \makecell[l]{ abusive $\rightarrow$ abusive                                               \\ hate
            $\rightarrow$ hateful                                                                                                                                                              \\ spam $\rightarrow$ spam \\ neutral $\rightarrow$ normal}

                                                & \texttt{X} $\rightarrow$
                                                \texttt{X} It was
                                                \texttt{{[}MASK{]}}                                                                                                       \\
            \hline
            MLMA fine-grained hostility En      & Twitter                              &
            5,549                               & 1,388                                & 1,735  & \makecell[l]{ abusive $\rightarrow$ abusive                                               \\ hate
            $\rightarrow$ hateful                                                                                                                                                              \\ offensive $\rightarrow$ offensive \\ disrespectful
            $\rightarrow$ disrespectful                                                                                                                                                     \\ fearful $\rightarrow$ fearful \\ neutral
            $\rightarrow$ normal}
            & \texttt{X} $\rightarrow$ \texttt{X} It was
            \texttt{{[}MASK{]}}                                                                                                                                                                      \\
            \hline
            OLID binary offensive En            & Twitter                              &
            10,592                              & 2,648                                & 860    & \makecell[l]{ offensive $\rightarrow$ offensive                                           \\
            neutral $\rightarrow$ normal}

                                                & \texttt{X} $\rightarrow$
                                                \texttt{X} It was
                                                \texttt{{[}MASK{]}}                                                                                                                                                             \\
                                                                                                            \hline
            SRW fine-grained abusive En                      & Twitter                              &
            6,504                               & 1,626                                & 2,033  & \makecell[l]{ sexist $\rightarrow$ sexism                                                 \\ racist
            $\rightarrow$ racism                                                                                                                                                            \\ neutral $\rightarrow$ normal}

                                                & \texttt{X} $\rightarrow$
                                                \texttt{X} It was
                                                \texttt{{[}MASK{]}}
            \\
            \hline
            Stormfront binary hate En           & Stormfront forum                     &
            6,849                               & 1,713                                & 2,141  & \makecell[l]{ hate $\rightarrow$ hate                                                     \\ neutral
            $\rightarrow$ normal}

                                                & \texttt{X}
                                                $\rightarrow$ \texttt{X} It was
                                                \texttt{{[}MASK{]}}\\
            \hline
            ToLD-Br fine-grained toxicity Pt-Br & Twitter                              &
            12,833                              & 3,209                                & 4,011  & \makecell[l]{ homophobic $\rightarrow$ LGBTQ+phobia                                         \\
            obscene $\rightarrow$ obscene                                                                                                                                                   \\ insulting $\rightarrow$ insult \\ racist
            $\rightarrow$ racism                                                                                                                                                            \\ sexist $\rightarrow$ misogyny \\ xenophobic $\rightarrow$
            xenophobia                                                                                                                                                                      \\ neutral $\rightarrow$ normal}
            & \texttt{X} $\rightarrow$ \texttt{X} It was \texttt{{[}MASK{]}}                                                                                                       \\
            \hline
            HASOC binary target En              & Twitter, Facebook                    &
            4,681                               & 1,171                                & 1,153  & \makecell[l]{ targeted $\rightarrow$ targeted                                             \\
            general $\rightarrow$ untargeted}

                                                & \texttt{X}
                                                $\rightarrow$ \texttt{X} It was
                                                \texttt{{[}MASK{]}}\\
            \hline
            AMI binary target En                & Twitter                              &
            1,428                               & 357                                  & 460    & \makecell[l]{ individual $\rightarrow$ active                                         \\
            group $\rightarrow$ passive}

                                                & \texttt{X} $\rightarrow$
                                                \texttt{X}
                                                It was targeted at
                                                \texttt{{[}MASK{]}}                                                                                                                                                   \\
            \hline
            HatEval binary target En            & Twitter                              &
            3,732                               & 933                                  & 1,318  & \makecell[l]{ individual $\rightarrow$ individual                                         \\
            group $\rightarrow$ group}

                                                & \texttt{X} $\rightarrow$
                                                \texttt{X}
                                                It was targeted at
                                                \texttt{{[}MASK{]}}                                                                                                                                                   \\
            \hline
            OLID fine-grained target En         & Twitter                              &
            3,100                               & 776                                  & 213    & \makecell[l]{ individual $\rightarrow$ individual                                         \\
            group $\rightarrow$ group                                                                                                                                                       \\ other $\rightarrow$ other}

                                                & \texttt{X} $\rightarrow$
                                                \texttt{X} It was targeted at
                                                \texttt{{[}MASK{]}}
        \end{tabular}    }
    \caption{
        Dataset statistics for each (dataset, label configuration, language)
        triple.
        From left to right we indicate the source platform of the dataset, the
        number of total train, validation and test samples, used verbalizers
        (\texttt{<predicted word>} $\rightarrow$ \texttt{<label>}) which also
        indicates the labels of a given dataset, and
        patterns (where \texttt{X} is the input sentence).
        We kept our PVPs simple, i.e., most labels are mapped 1-to-1 to the same
        word, and we defined only two patterns.
        Note that we also used English PVPs for non-English datasets, since it
        was shown to perform well \cite{zhao-schutze-2021-discrete}.
        Since different datasets often name the negative abuse class
         differently (e.g.
        no-hate, not-offensive, normal, etc.), we unified them by using the
        frequent \emph{normal} label name.
        Additionally, similarly defined but differently named labels, such as
        hate and hateful or sexism and misogyny, are united by using the same
        verbalizers for them.
    }
    \label{tab:datasets}
\end{table*}

\begin{table*}[t]
    \centering
    \newcommand\myrow{
    \csvcolvi & \perc{\csvcolvii}\std{\csvcolviii} & \perc{\csvcolix}\std{\csvcolx}
    & \perc{\csvcolxi}\std{\csvcolxii} & \perc{\csvcolxiii}\std{\csvcolxiv}
    }
    \begin{subtable}{0.43\linewidth}
        \resizebox{\textwidth}{!}{        
\begin{tabular}{r|rrrr}
                       & hate & offen. & profa. &  avg. \\
            \hline
            \hline
            \multicolumn{5}{c}{1-shot} \\
            \hline
            \csvreader[range = 22-23, late after line = \\]{results_hasoc_en-abusive.csv}{}{
                \myrow
            }
            \hline
            \multicolumn{5}{c}{4-shot} \\
            \hline
            \csvreader[range = 40-45, late after line = \\]{results_hasoc_en-abusive.csv}{}{
                \myrow
            }
            \hline
            \multicolumn{5}{c}{8-shot} \\
            \hline
            \csvreader[range = 55-56, late after line = \\]{results_hasoc_en-abusive.csv}{}{
                \myrow
            }
            \hline
            \multicolumn{5}{c}{16-shot} \\
            \hline
            \csvreader[range = 59-60, late after line = \\]{results_hasoc_en-abusive.csv}{}{
                \myrow
            }
            \hline
            \multicolumn{5}{c}{32-shot} \\
            \hline
            \csvreader[range = 63-64, late after line = \\]{results_hasoc_en-abusive.csv}{}{
                \myrow
            }
            \hline
            \multicolumn{5}{c}{64-shot} \\
            \hline
            \csvreader[range = 67-68, late after line = \\]{results_hasoc_en-abusive.csv}{}{
                \myrow
            }
    \end{tabular}        }
        \caption{HASOC fine-grained abusive En}
    \end{subtable}
    \hfill
    \vspace{0.2cm}
    \renewcommand\myrow{
    \csvcolvi & \perc{\csvcolvii}\std{\csvcolviii} & \perc{\csvcolix}\std{\csvcolx}
    & \perc{\csvcolxi}\std{\csvcolxii} & \perc{\csvcolxiii}\std{\csvcolxiv}
    }
    \begin{subtable}{0.43\linewidth}
        \resizebox{\textwidth}{!}{        \begin{tabular}{r|rrrr}
                                  & hate & offen. & profa. &  avg. \\
    \hline
    \hline
    \multicolumn{5}{c}{1-shot} \\
\end{tabular}        }
        \caption{HASOC fine-grained abusive De}
    \end{subtable}
    \hfill
    \vspace{0.2cm}
    \renewcommand\myrow{
    \csvcolvi & \perc{\csvcolvii}\std{\csvcolviii} & \perc{\csvcolix}\std{\csvcolx}
    & \perc{\csvcolxi}\std{\csvcolxii} & \perc{\csvcolxiii}\std{\csvcolxiv}
    }
    \begin{subtable}{0.43\linewidth}
        \resizebox{\textwidth}{!}{\begin{tabular}{r|rrrr}
                                & hate & offen. & profa. &  avg. \\
    \hline
    \hline
    \multicolumn{5}{c}{1-shot} \\
\end{tabular}        }
        \caption{HASOC fine-grained abusive Hi}
    \end{subtable}
    \hfill
    \vspace{0.2cm}
    \renewcommand\myrow{
    \csvcolvi & \perc{\csvcolvii}\std{\csvcolviii} & \perc{\csvcolix}\std{\csvcolx}
    & \perc{\csvcolxiii}\std{\csvcolxiv} & \perc{\csvcolxi}\std{\csvcolxii} &
    \perc{\csvcolxv}\std{\csvcolxvi}
    }
    \begin{subtable}{0.43\linewidth}
        \resizebox{\textwidth}{!}{        \begin{tabular}{r|rrrrr}
                        & abus. & insult & profa. & normal & avg. \\
    \hline
    \hline
    \multicolumn{6}{c}{1-shot} \\
\end{tabular}
        }
        \caption{GermEval fine-grained offensive De}
    \end{subtable}
    \hfill
    \vspace{0.2cm}

    \renewcommand\myrow{
    \csvcolvi & \perc{\csvcolxiii}\std{\csvcolxiv} &
    \perc{\csvcolxix}\std{\csvcolxx} & \perc{\csvcolix}\std{\csvcolx} &
    \perc{\csvcolxxi}\std{\csvcolxxii} & \perc{\csvcolvii}\std{\csvcolviii} &
    \perc{\csvcolxvii}\std{\csvcolxviii} & \perc{\csvcolxv}\std{\csvcolxvi} &
    \perc{\csvcolxi}\std{\csvcolxii}
    }
    \begin{subtable}{0.77\linewidth}
        \resizebox{\textwidth}{!}{        \begin{tabular}{r|rrrrrrrr}
                       & misogyny & racism & insult & xenophobia & LGBTQ+hobia & obscene & normal &  avg. \\
    \hline
    \hline
    \multicolumn{9}{c}{1-shot} \\
    \hline
    \csvreader[range = 15-16, late after line = \\]{results_told-br_fine_grained.csv}{}{
        \myrow
    }
    \hline
    \multicolumn{9}{c}{4-shot} \\
    \hline
    \csvreader[range = 35-40, late after line = \\]{results_told-br_fine_grained.csv}{}{
        \myrow
    }
    \hline
    \multicolumn{9}{c}{8-shot} \\
    \hline
    \csvreader[range = 50-51, late after line = \\]{results_told-br_fine_grained.csv}{}{
        \myrow
    }
\end{tabular}
        }
        \caption{ToLD-Br fine-grained toxic Pt-Br}
    \end{subtable}
        \addtocounter{table}{-1}
\end{table*}

\begin{table*}
\centering
    \newcommand\myrow{
    \csvcolvi & \perc{\csvcolvii}\std{\csvcolviii} &
    \perc{\csvcolix}\std{\csvcolx} & \perc{\csvcolxi}\std{\csvcolxii} &
    \perc{\csvcolxiii}\std{\csvcolxiv}
    }
    \begin{subtable}{0.47\linewidth}
        \setcounter{subtable}{5}
        \resizebox{\textwidth}{!}{        \begin{tabular}{r|rrrr}
                        & group & indivi. & other &  avg. \\
    \hline
    \hline
    \multicolumn{5}{c}{1-shot} \\
    \hline
    \csvreader[range = 24-25, late after line = \\]{results_olid_fine_grained_target.csv}{}{
        \myrow
    }
    \hline
    \multicolumn{5}{c}{4-shot} \\
    \hline
    \csvreader[range = 43-48, late after line = \\]{results_olid_fine_grained_target.csv}{}{
        \myrow
    }
    \hline
    \multicolumn{5}{c}{8-shot} \\
    \hline
    \csvreader[range = 60-61, late after line = \\]{results_olid_fine_grained_target.csv}{}{
        \myrow
    }
\end{tabular}
        }
        \caption{OLID fine-grained target En}
    \end{subtable}
    \hfill
    \vspace{0.2cm}
    \renewcommand\myrow{
    \csvcolvi & \perc{\csvcolvii}\std{\csvcolviii} &
    \perc{\csvcolxi}\std{\csvcolxii} & \perc{\csvcolix}\std{\csvcolx}
    }
    \begin{subtable}{0.37\linewidth}
        \resizebox{\textwidth}{!}{        \begin{tabular}{r|rrr}
                       & hate & normal &  avg. \\
    \hline
    \hline
    \multicolumn{4}{c}{1-shot} \\
    \hline
    \csvreader[range = 23-24, late after line = \\]{results_stormfront_binary.csv}{}{
        \myrow
    }
    \hline
    \multicolumn{4}{c}{4-shot} \\
    \hline
    \csvreader[range = 43-48, late after line = \\]{results_stormfront_binary.csv}{}{
        \myrow
    }
    \hline
    \multicolumn{4}{c}{8-shot} \\
    \hline
    \csvreader[range = 59-60, late after line = \\]{results_stormfront_binary.csv}{}{
        \myrow
    }
    \hline
    \multicolumn{4}{c}{16-shot} \\
    \hline
    \csvreader[range = 64-65, late after line = \\]{results_stormfront_binary.csv}{}{
        \myrow
    }
    \hline
    \multicolumn{4}{c}{32-shot} \\
    \hline
    \csvreader[range = 69-70, late after line = \\]{results_stormfront_binary.csv}{}{
        \myrow
    }
    \hline
    \multicolumn{4}{c}{64-shot} \\
    \hline
    \csvreader[range = 74-75, late after line = \\]{results_stormfront_binary.csv}{}{
        \myrow
    }
\end{tabular}
        }
        \caption{Stormfront binary hate En}
    \end{subtable}
    \hfill
    \vspace{0.2cm}
    \renewcommand\myrow{
    \csvcolvi & \perc{\csvcolvii}\std{\csvcolviii} &
    \perc{\csvcolxi}\std{\csvcolxii} & \perc{\csvcolix}\std{\csvcolx}
    }
    \begin{subtable}{0.39\linewidth}
        \resizebox{\textwidth}{!}{        \begin{tabular}{r|rrr}
                       & hateful & normal &  avg. \\
    \hline
    \hline
    \multicolumn{4}{c}{1-shot} \\
\end{tabular}
        }
        \caption{HatEval binary hate En}
    \end{subtable}
    \hfill
    \vspace{0.2cm}
    \renewcommand\myrow{
    \csvcolvi & \perc{\csvcolvii}\std{\csvcolviii} &
    \perc{\csvcolxi}\std{\csvcolxii} & \perc{\csvcolix}\std{\csvcolx}
    }
    \begin{subtable}{0.40\linewidth}
        \resizebox{\textwidth}{!}{        \begin{tabular}{r|rrr}
                       & hateful & normal &  avg. \\
    \hline
    \hline
    \multicolumn{4}{c}{1-shot} \\
\end{tabular}
        }
        \caption{HatEval binary hate Es}
    \end{subtable}
    \hfill
    \vspace{0.2cm}
\addtocounter{table}{-1}
\end{table*}

\begin{table*}
\centering
    \newcommand\myrow{
    \csvcolvi & \perc{\csvcolix}\std{\csvcolx} & \perc{\csvcolvii}\std{\csvcolviii} &
    \perc{\csvcolxi}\std{\csvcolxii}
    }
    \begin{subtable}{0.40\linewidth}
    \setcounter{subtable}{9}
        \resizebox{\textwidth}{!}{        \begin{tabular}{r|rrr}
                          & offensive & normal &  avg. \\
    \hline
    \hline
    \multicolumn{4}{c}{1-shot} \\
\end{tabular}
        }
        \caption{OLID binary offensive En}
    \end{subtable}
    \hfill
    \vspace{0.2cm}
    \renewcommand\myrow{
    \csvcolvi & \perc{\csvcolvii}\std{\csvcolviii} & \perc{\csvcolix}\std{\csvcolx} &
    \perc{\csvcolxi}\std{\csvcolxii}
    }
    \begin{subtable}{0.40\linewidth}
        \resizebox{\textwidth}{!}{        \begin{tabular}{r|rrr}
                          & offensive & normal &  avg. \\
    \hline
    \hline
    \multicolumn{4}{c}{1-shot} \\
\end{tabular}
        }
        \caption{GermEval binary offensive De}
    \end{subtable}
    \hfill
    \vspace{0.2cm}
    \renewcommand\myrow{
    \csvcolvi & \perc{\csvcolix}\std{\csvcolx} &
    \perc{\csvcolxi}\std{\csvcolxii} & \perc{\csvcolvii}\std{\csvcolviii}
    }
    \begin{subtable}{0.43\linewidth}
        \resizebox{\textwidth}{!}{        \begin{tabular}{r|rrr}
                              & misogyny & normal &  avg. \\
    \hline
    \hline
    \multicolumn{4}{c}{1-shot} \\
\end{tabular}
        }
        \caption{AMI binary sexism En}
    \end{subtable}
    \hfill
    \vspace{0.2cm}
    \renewcommand\myrow{
    \csvcolvi & \perc{\csvcolix}\std{\csvcolx} &
    \perc{\csvcolxi}\std{\csvcolxii} & \perc{\csvcolvii}\std{\csvcolviii}
    }
    \begin{subtable}{0.40\linewidth}
        \resizebox{\textwidth}{!}{        \begin{tabular}{r|rrr}
                              & misogyny & normal &  avg. \\
    \hline
    \hline
    \multicolumn{4}{c}{1-shot} \\
    \hline
    \csvreader[range = 30-31, late after line = \\]{results_ami_it-misogyny.csv}{}{
        \myrow
    }
    \hline
    \multicolumn{4}{c}{4-shot} \\
    \hline
        \csvreader[range = 55-60, late after line = \\]{results_ami_it-misogyny.csv}{}{
        \myrow
    }
    \hline
    \multicolumn{4}{c}{8-shot} \\
    \hline
    \csvreader[range = 75-76, late after line = \\]{results_ami_it-misogyny.csv}{}{
        \myrow
    }
    \hline
    \multicolumn{4}{c}{16-shot} \\
    \hline
    \csvreader[range = 79-80, late after line = \\]{results_ami_it-misogyny.csv}{}{
        \myrow
    }
    \hline
    \multicolumn{4}{c}{32-shot} \\
    \hline
    \csvreader[range = 83-84, late after line = \\]{results_ami_it-misogyny.csv}{}{
        \myrow
    }
    \hline
    \multicolumn{4}{c}{64-shot} \\
    \hline
    \csvreader[range = 87-88, late after line = \\]{results_ami_it-misogyny.csv}{}{
        \myrow
    }
\end{tabular}
        }
        \caption{AMI binary misogyny It}
    \end{subtable}
    \hfill
    \vspace{0.2cm}
    \caption{
        Per label and macro averaged $F_1$ scores for each target dataset.
        \emph{LM-base}, \emph{MLM}, \emph{MTL} and \emph{Fusion} rows indicate
        the model trained on the target task only,
        the masked language modeling, multitask learning and adapter fusion
        baselines,
        \emph{MDT} our proposed approach,
        while
        \emph{MDT-abl.} refers to the ablation studies where external only labels are
        removed from the external datasets.
                                      }
    \label{tab:all_results}
\end{table*}

\end{document}